\documentclass[10pt,twocolumn,letterpaper]{article}
\usepackage[final]{cvpr}

\usepackage{cite}
\usepackage{graphicx}
\usepackage{amsmath}
\usepackage{bm}
\DeclareMathOperator*{\argmax}{argmax}
\usepackage{amssymb}
\usepackage{booktabs}
\usepackage[utf8]{inputenc}
\usepackage{mathtools}

\usepackage[dvipsnames]{xcolor}
\usepackage{kotex}
\usepackage{multirow}
\usepackage{booktabs}
\usepackage{makecell}
\usepackage[export]{adjustbox}
\usepackage{upquote}
\usepackage{stackengine} 
\usepackage{scalerel}
\usepackage{soul}
\usepackage{tabularx}
\makeatletter
\@namedef{ver@everyshi.sty}{}
\makeatother
\usepackage{tikz}
\usepackage{algorithm} 
\usepackage{algpseudocode} 
\usepackage{eqparbox}
\usepackage{colortbl}
\usepackage[hang,flushmargin]{footmisc} 
\usepackage[hypcap=true]{caption}
\newcolumntype{Y}{>{\centering\arraybackslash}X}

\usepackage[colorlinks,citecolor=blue,pagebackref,breaklinks=true,bookmarks=false]{hyperref}

\usepackage[capitalize]{cleveref} %
\crefname{section}{Sec.}{Secs.}
\Crefname{section}{Section}{Sections}
\Crefname{table}{Table}{Tables}
\crefname{table}{Table}{Tabs.}
\creflabelformat{equation}{#2\textup{#1}#3}

\def\cvprPaperID{5956}%
\def\confName{CVPR}
\def\confYear{2022}

\belowdisplayskip=\belowdisplayshortskip
\abovedisplayskip=\abovedisplayshortskip
\begin{document}

\newcommand{\Eq}[1]  {Eq.\ (#1)}
\newcommand{\Eqs}[1] {Eqs.\ (#1)}
\newcommand{\Fig}[1] {Fig.\ #1}
\newcommand{\Figs}[1]{Figs.\ #1}
\newcommand{\Tbl}[1]  {Table\ #1}
\newcommand{\Tbls}[1] {Tables\ #1}
\newcommand{\Sec}[1] {Sec.\ #1}
\newcommand{\SSec}[1] {Sec.\ #1}
\newcommand{\Secs}[1] {Secs.\ #1}
\newcommand{\Alg}[1] {Alg.\ #1}
\newcommand{\Etal}   {{\textit{et al.}}}

\newcommand{\setone}[1] {\left\{ #1 \right\}} %
\newcommand{\settwo}[2] {\left\{ #1 \mid #2 \right\}} %

\newcommand{\todo}[1]{{\textcolor{red}{#1}}}
\newcommand{\son}[1]{{\textcolor{magenta}{hyeongseok: #1}}}
\newcommand{\jy}[1]{{\textbf{\textcolor{MidnightBlue}{[JY] }}\textcolor{MidnightBlue}{#1}}}
\newcommand{\sean}[1]{{\textcolor{green}{sean: #1}}}
\newcommand{\sunghyun}[1]{{\textcolor[rgb]{0.6,0.0,0.6}{sunghyun: #1}}}
\newcommand{\mh}[1]{{\textcolor[rgb]{0.6,0.0,0.1}{mhlee: #1}}}
\newcommand{\change}[1]{{\color{black}#1}}
\newcommand{\changed}[1]{{\color{blue}#1}}
\newcommand{\bb}[1]{\textbf{\textit{#1}}}

\renewcommand{\topfraction}{0.95}
\setcounter{bottomnumber}{1}
\renewcommand{\bottomfraction}{0.95}
\setcounter{totalnumber}{3}
\renewcommand{\textfraction}{0.05}
\renewcommand{\floatpagefraction}{0.95}
\setcounter{dbltopnumber}{2}
\renewcommand{\dbltopfraction}{0.95}
\renewcommand{\dblfloatpagefraction}{0.95}

\newcommand{\Net}[1]{#1}
\newcommand{\Loss}[1]{$\mathcal{L}_{#1}$}
\newcommand{\cm}{\checkmark}
\newcommand{\ts}{\textsuperscript}
\newcommand\oast{\stackMath\mathbin{\stackinset{c}{0ex}{c}{0ex}{\ast}{\bigcirc}}}

\makeatletter
\newcommand{\StatexIndent}[1][3]{%
  \setlength\@tempdima{\algorithmicindent}%
  \Statex\hskip\dimexpr#1\@tempdima\relax}
\makeatother

\newdimen{\algindent}
\setlength\algindent{1.5em}
\algnewcommand\LeftComment[2]{%
\hspace{#1\algindent}$\triangleright$ \eqparbox{COMMENT}{#2} \hfill %
}
\newcommand*\pct{\protect\scalebox{0.9}{\%}}
\newcommand*\smalleq{\protect\scalebox{0.9}{=}}
\newcommand*\MAES{\protect\scalebox{0.6}{${(\!\times\!10^{\texttt{-}1})}$}}
\def\nespace{\hskip\fontdimen2\font\relax}
\newcommand{\midsepremove}{\aboverulesep = 0mm \belowrulesep = 0mm}
\midsepremove
\newcommand{\midsepdefault}{\aboverulesep = 0.605mm \belowrulesep = 0.984mm}
\midsepdefault

\newcommand*{\bbox}[1]{%
   \begingroup
   \setlength{\fboxrule}{0.1pt}%
   \setlength{\fboxsep}{-0.1pt}%
   \fbox{\mathsurround=0pt{$#1$}}%
   \endgroup
}

\definecolor{lightlightgray}{gray}{0.96}
\definecolor{vividyellow}{RGB}{255, 152, 0}
\definecolor{emeraldgreen}{RGB}{42, 148, 0}
\newcommand\blfootnote[1]{%
  \begingroup
  \renewcommand\thefootnote{}\footnote{#1}%
  \addtocounter{footnote}{-1}%
  \endgroup
}
\title{Reference-based Video Super-Resolution Using Multi-Camera Video Triplets}
\author{
    Junyong Lee
    \qquad
    Myeonghee Lee
    \qquad
    Sunghyun Cho
    \qquad
    Seungyong Lee\\
    POSTECH\\
    {\tt\small
    \{junyonglee, myeonghee, s.cho, leesy\}@postech.ac.kr}
}
\maketitle
\begin{abstract}
We propose the first reference-based video super-resolution (RefVSR) approach that utilizes reference videos for high-fidelity results. We focus on RefVSR in a triple-camera setting, where we aim at super-resolving a low-resolution ultra-wide video utilizing wide-angle and telephoto videos. We introduce the first RefVSR network that recurrently aligns and propagates temporal reference features fused with features extracted from low-resolution frames. To facilitate the fusion and propagation of temporal reference features, we propose a propagative temporal fusion module. For learning and evaluation of our network, we present the first RefVSR dataset consisting of triplets of ultra-wide, wide-angle, and telephoto videos concurrently taken from triple cameras of a smartphone. We also propose a two-stage training strategy fully utilizing video triplets in the proposed dataset for real-world 4$\times$ video super-resolution. We extensively evaluate our method, and the result shows the state-of-the-art performance in 4$\times$ super-resolution.
\end{abstract} 
\begin{figure}[t]
\begin{center}
\includegraphics [width=1.0\linewidth] {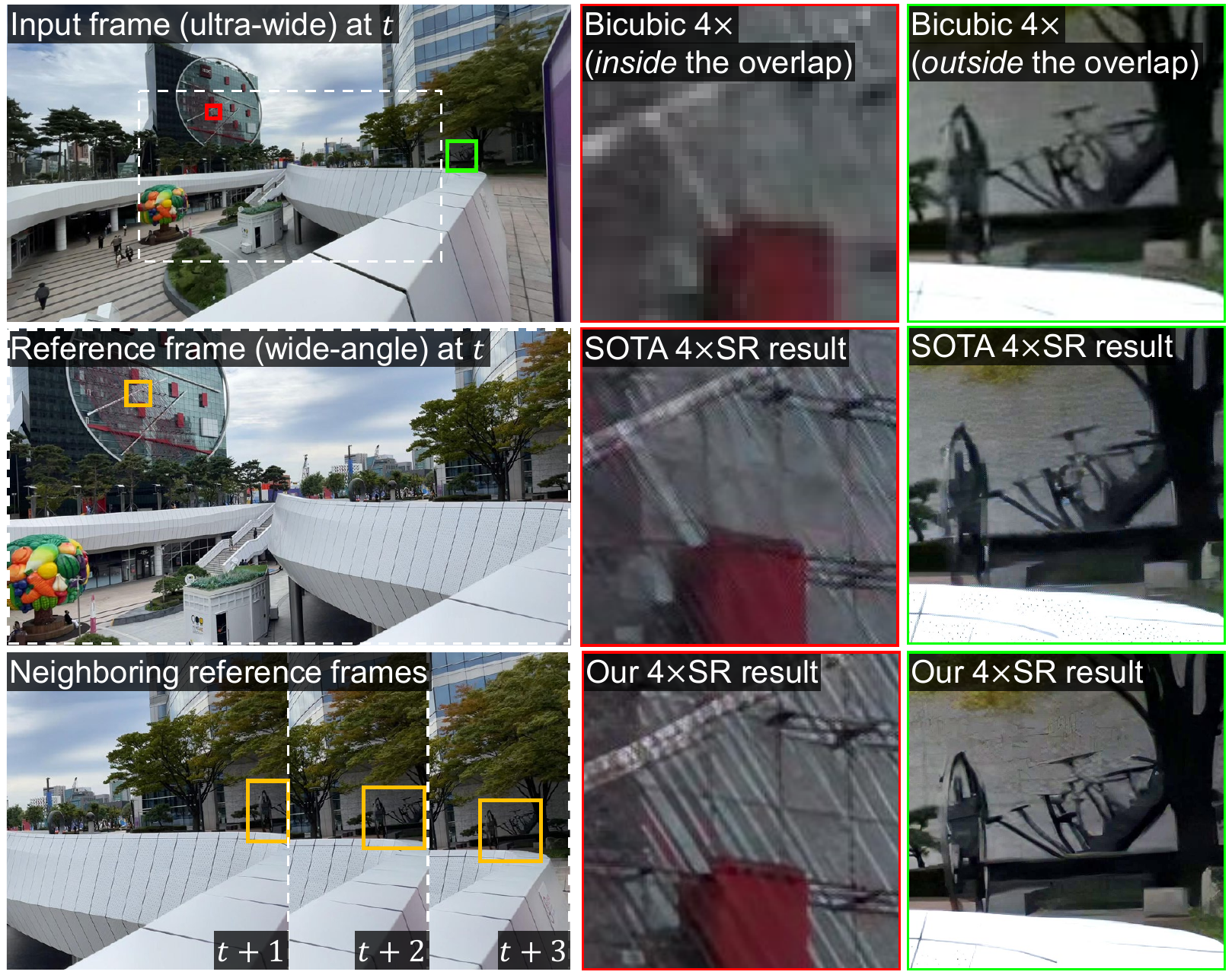}
\end{center}
\vspace{-0.6cm}
\caption{Comparison on 8K 4$\times$SR video results from a real HD video between state-of-the-art (SOTA) RefSR approach~\cite{wang2021DCSR} and the proposed RefVSR approach.
Our method learns to super-resolve an LR video by utilizing relevant high-quality patches of \textcolor{vividyellow}{reference frames} and robustly recovers sharp textures of both \textcolor{red}{inside} and \textcolor{emeraldgreen}{outside} the overlapped FoV between the input ultra-wide and reference wide-angle frames (white dashed box).}
\label{fig:teaser}
\vspace{-8pt}
\end{figure}

\section{Introduction} \label{sec:Intorduction}
\blfootnote{Code and dataset: \url{https://github.com/codeslake/RefVSR}}
Recent mobile devices such as Apple iPhone or Samsung Galaxy series are manufactured with at least two or three asymmetric multi-cameras typically having different but fixed focal lengths.
In a triple camera setting, each ultra-wide, wide-angle, and telephoto camera has a different field of view (FoV) and optical zoom factor.
One advantage of such configuration is that, compared to an ultra-wide camera, a wide-angle camera captures a subject with more details and higher resolution, and the advantage escalates even further with a telephoto camera.
A question naturally follows is why not leverage higher-resolution frames of a camera with a longer focal length to improve the resolution of frames of a camera with a short focal length.

Utilizing a reference (Ref) image to reconstruct a high-resolution (HR) image from a low-resolution (LR) image has been widely studied in previous reference-based image super-resolution (RefSR) approaches~\cite{Boominathan2014Improving, zheng2017patchamatchSR, zheng2018crossnet, Zhang2019ImageSB, yang2020TTSR, Xie2020FeatureRM, shim2020SSEN, wang2021DCSR}.
However, 
it has not been explored yet to utilize a Ref video for video super-resolution (VSR).
In this paper, we expand the RefSR to the VSR task and introduce reference-based video super-resolution (RefVSR) that can be applied for videos captured in an asymmetric multi-camera setting.

RefVSR inherits objectives of both RefSR and VSR tasks and utilizes a Ref video for reconstructing an HR video from an LR video.
Applying RefVSR for a video captured in an asymmetric multi-camera setting requires consideration of the unique relationship between LR and Ref frames in multi-camera videos.
In the setting, a pair of LR and Ref frames at each time step shares almost the same content in their overlapped FoV (top and middle rows of the leftmost column in \cref{fig:teaser}).
Moreover, 
as a video exhibits a motion, neighboring Ref frames might contain high-quality contents useful for recovering the outside the overlapped FoV (the bottom row of the leftmost column in \cref{fig:teaser}).%

For successful RefVSR in an asymmetric multi-camera setting, we take advantage of temporal Ref frames in reconstructing regions both inside and outside the overlapped FoV.
In previous RefSR approaches~\cite{Zhang2019ImageSB,yang2020TTSR,Xie2020FeatureRM,wang2021DCSR}, global matching has been a common choice for establishing non-local correspondence between a pair of LR and Ref images.
However, given a pair of LR and Ref video sequences, it is not straightforward to directly apply global matching between an LR frame and {\em multiple} Ref frames.
To utilize as many frames as possible in the global matching for large real-world videos (\eg, HD videos),
we need a framework capable of managing Ref frames in a memory-efficient way.

We propose the first end-to-end learning-based RefVSR network that can generally be applied for super-resolving an LR video using a Ref video.
Our network adopts a bidirectional recurrent pipeline~\cite{huang2015brcn, huang2018brcn, chan2021basicvsr} to recurrently align and propagate Ref features that are fused with the features of LR frames.
Our network is efficient in terms of computation and memory consumption because the global matching needed for aligning Ref features is performed only between a pair of LR and corresponding Ref frames at each time step.
Still, our network is capable of utilizing temporal Ref frames, as the aligned Ref features are continuously fused and propagated in the pipeline.

As a key component for managing Ref features in the pipeline,
we propose a propagative temporal fusion module
that fuses and propagates only well-matched Ref features.
The module leverages the matching confidence computed during the global matching between LR and Ref features as the guidance to determine well-matched Ref features to be fused and propagated.
The module also accumulates the matching confidence throughout the pipeline and uses the accumulated value as the guidance when fusing the propagated temporal Ref features.

To train and validate our model, 
we present the first RefVSR dataset consisting of 161 video triplets of ultra-wide, wide-angle, and telephoto videos simultaneously captured with triple cameras of a smartphone.
Wide-angle and telephoto videos have the same size as ultra-wide videos but their resolutions are 2$\times$ and 4$\times$ the resolution of ultra-wide videos, respectively.
With the RefVSR dataset, we train our network to super-resolve an ultra-wide video 4$\times$ to produce an 8K video with the same resolution as a telephoto video.
To this end, we propose a two-stage training strategy that fully utilizes video triplets in the proposed dataset. %
We show that, with our training strategy, our network can successfully learn super-resolution of a real-world HD video and produce a high-fidelity 8K video.

\vspace{4pt}
To summarize, our contributions include:
\begin{itemize}
    \vspace{-4pt}
    \setlength\itemsep{-0.1em}
    \item the first RefVSR framework with the focus on videos recorded in an asymmetric multi-camera setting,
    \item the propagative temporal fusion module that effectively fuses and propagates temporal Ref features,
    \item the RealMCVSR dataset, which is the first dataset for the RefVSR task, and
    \item the two-stage training strategy fully utilizing video triplets for real-world 4$\times$VSR.
\end{itemize}

\section{Related Work}
\paragraph{Reference-based Super-Resolution (RefSR)}
Previous RefSR approaches~\cite{Boominathan2014Improving, zheng2017patchamatchSR, zheng2018crossnet, Zhang2019ImageSB, yang2020TTSR, Xie2020FeatureRM, shim2020SSEN, wang2021DCSR}
have focused on establishing non-local correspondence between LR and Ref features.
For establishing correspondence, either offset-based matching (optical flow~\cite{zheng2018crossnet} and deformable convolution~\cite{shim2020SSEN}) or patch-based matching (patch-match~\cite{Boominathan2014Improving, zheng2017patchamatchSR, Barnes:2009:PAR, Zhang2019ImageSB}, learnable patch-match~\cite{yang2020TTSR, Xie2020FeatureRM}, learnable patch-match with affine correction~\cite{wang2021DCSR}) are employed.

\vspace{-12pt}
\paragraph{Video Super-Resolution (VSR)}
Previous VSR methods have focused on how to effectively utilize highly related but unaligned LR frames in a video sequence.
With respect to how LR frames in video sequences are handled by a model,
previous VSR approaches can be categorized into either sliding window-based~\cite{caballero2017Real, wang2019edvr, li2019Fast, tian2020tdan, li2020mucan} or recurrent framework-based~\cite{huang2015brcn, huang2018brcn, Sajjadi2018frvsr, Isobe2020VideoSW, chan2021basicvsr} approaches.
For handling unaligned LR frames, warping using optical flow~\cite{caballero2017Real, Sajjadi2018frvsr, chan2021basicvsr}, patch-based correlation~\cite{li2020mucan}, and deformable convolution~\cite{wang2019edvr, tian2020tdan} have been employed.

The aforementioned previous studies in RefSR and VSR have developed various components. %
In this paper, to match and align Ref features to an LR frame, we adopt the learnable patch-match-based reference alignment module~\cite{yang2020TTSR, Xie2020FeatureRM, wang2021DCSR}.
To handle video sequences, we adopt a bidirectional recurrent framework~\cite{huang2015brcn, huang2018brcn, chan2021basicvsr}.
However, for RefVSR, we modify the components to handle both LR and Ref videos.
We also equip our network with the propagative temporal fusion module, designed to effectively and efficiently exploit temporal Ref features in reconstructing HR frames.

Recently, a RefVSR method~\cite{Zhao2021EFENet}, in which only the first frame of an HR video is used as a reference to super-resolve an LR video downsampled from the HR video, has been concurrently proposed alongside our work.
However, to the best of our knowledge, ours is the first RefVSR framework that utilizes multiple frames in a Ref video for super-resolving a real-world LR video.

\section{Multi-Camera Video Super-Resolution}
\subsection{Framework Overview}
\label{sec:framework}
\begin{figure}[t]
\begin{center}
\includegraphics [width=1.00\linewidth] {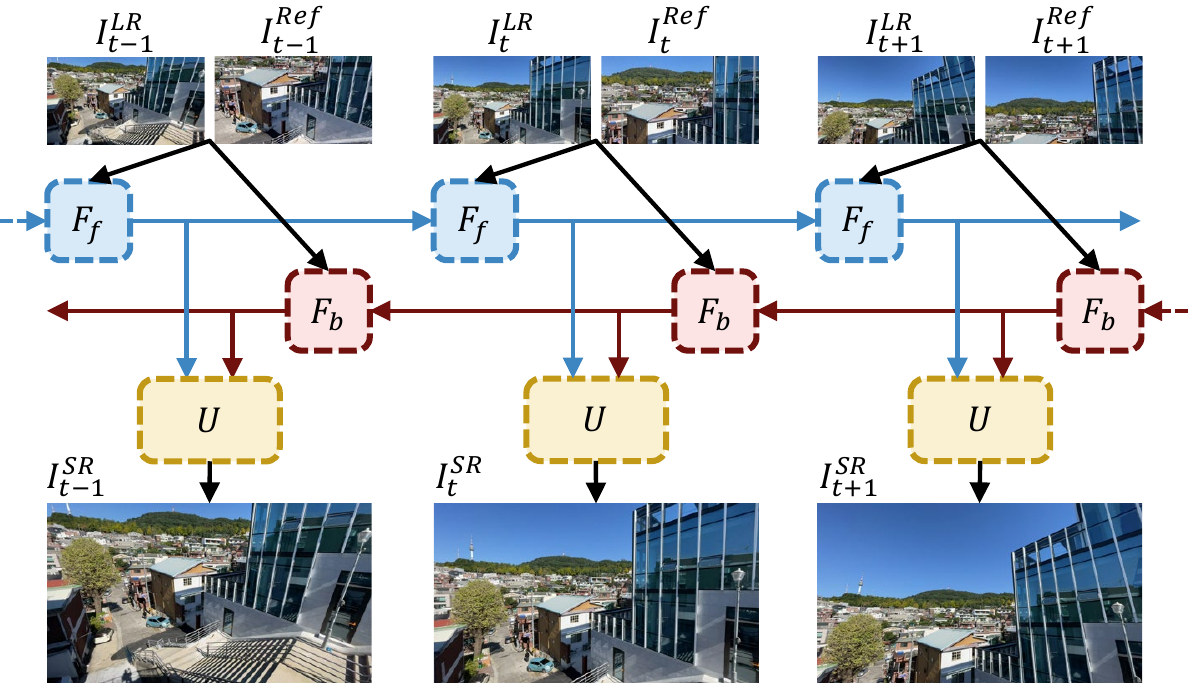}
\end{center}
\vspace{-0.6cm}
  \caption{Overview of our RefVSR framework.}
\label{fig:framework}
\vspace{-12pt}
\end{figure}

\begin{figure}[t]
\begin{center}
\includegraphics [width=1.00\linewidth] {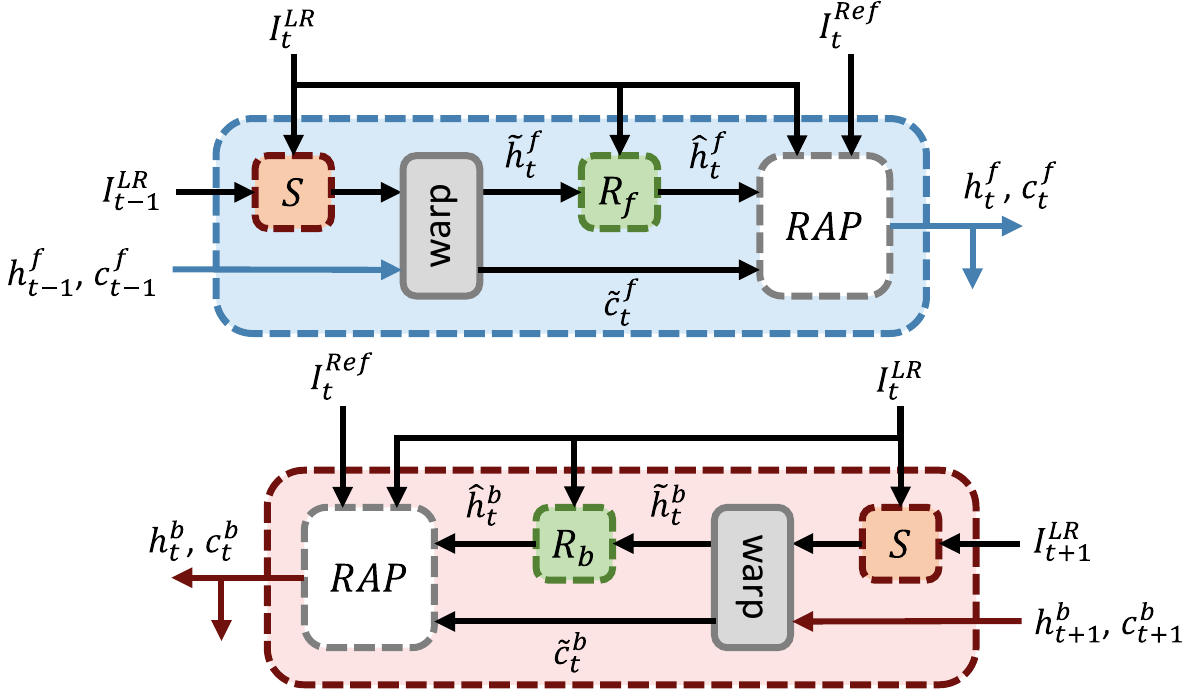}
\end{center}
\vspace{-0.6cm}
  \caption{Forward (top) and backward (bottom) recurrent cells.}
\label{fig:cell}
\vspace{-12pt}
\end{figure}

\cref{fig:framework} shows an overview of the proposed network,
which can generally be applied to a RefVSR task for super-resolving an LR video utilizing a Ref video.
Our network follows a typical bidirectional propagation scheme~\cite{huang2015brcn, chan2021basicvsr}, consisting of bidirectional recurrent cells $F_f$ and $F_b$, where the subscripts $f$ and $b$ indicate forward and backward propagation branches, respectively (\cref{fig:cell}).
Our network is distinguished from previous ones in additional inputs, intermediate features, and modules to utilize a Ref video sequence.

Specifically, for a time step $t$, each recurrent cell $F_f$ or $F_b$ takes not only low-resolution LR frames $I^{\smash{LR}}_{t\pm1}$ at the previous time step and $I^{\smash{LR}}_{t}$ at the current time step, 
but also a Ref frame $I^{\smash{Ref}}_{t}$ at the current time step.
Each cell is also recurrently fed with aggregated LR and Ref features $h^{\smash{\{f,b\}}}_{t\pm1}$ and accumulated confidence maps $c^{\smash{\{f,b\}}}_{t\pm1}$ propagated from the previous time step.
Here, the accumulated confidence maps are utilized for fusing well-matched Ref features later in each recurrent cell.
Finally, each recurrent cell propagates the resulting features $h^{\smash{\{f,b\}}}_{t}$ and the accumulated matching confidences $c^{\smash{\{f,b\}}}_{t}$ to the next cell.
Formally, we have:
\begin{equation}
\begin{split}
    \{h^f_t, c^{f}_{t}\}&\!=\!F_f(I^{LR}_{t-1},\, I^{LR}_t,\, I^{Ref}_t,\, h^f_{t-1},\, c^{f}_{t-1}),\\
    \{h^b_t, c^{b}_{t}\}&\!=\!F_b(I^{LR}_{t+1},\, I^{LR}_t,\, I^{Ref}_t,\, h^b_{t+1},\, c^{b}_{t+1}).
\end{split}
\end{equation}

For reconstructing an SR result $I^{SR}_{t}$, the upsampling module $U$ first takes the intermediate features $h^{\smash{\{f,b\}}}_{t}$ and accumulated matching confidences $c^{\smash{\{f,b\}}}_{t}$ of both forward and backward branches.
Then, the features are aggregated and upsampled with multiple convolution and pixel-shuffle~\cite{shi2016pixelshuffel} layers to produce $I^{SR}_{t}$.
Mathematically, we have:
\begin{equation}
\begin{gathered}[b]
    I^{SR}_{t}\!=\!U(h^{f}_{t},\, h^{b}_{t},\, c^{f}_{t},\, c^{b}_{t}).
\end{gathered}
\end{equation}

For the upsampling module $U$ to accurately reconstruct $I^{SR}_t$, 
the intermediate features $h^{\smash{\{f,b\}}}_{t}$
should contain details integrated from both LR and Ref frames in a video sequence.
To this end, each recurrent cell $F_f$ and $F_b$ performs inter-frame alignment between the previous and current LR input frames, then aggregates and propagates the features (\cref{sec:cell}).
To exploit multiple Ref frames, each recurrent cell aligns the current Ref features to the current LR frame and fuses the aligned Ref features to the aggregated features of the previous Ref, LR, and current LR frames using a reference alignment and propagation module (\cref{sec:RAP}).
In this way, features of temporally distant LR input and Ref frames can be recurrently integrated and propagated.

\subsection{Bidirectional Recurrent Cells}
\label{sec:cell}

In each recurrent cell $F_f$ and $F_b$ (\cref{fig:cell}), we first use a flow estimation network $S$~\cite{ranjan2017spynet} to estimate the optical flow between the LR frame $I_t^{\smash{LR}}$ at the current time step and $I_{t\pm1}^{\smash{LR}}$ at the previous time step to align propagated features $h_{t\pm1}^{\smash{\{f,b\}}}$ to $I_t^{\smash{LR}}$.
Then, using a residual block $R$, we aggregate an LR frame $I^{\smash{LR}}_{t}$ into the aligned features to obtain \emph{temporally aggregated features} $\widehat{h}^{\smash{\{f,b\}}}_{t}$.
Specifically, we have:
\begin{equation}
\begin{split}
    w^{\{f,b\}}_{t}&\!=\!S(I^{LR}_{t},\, I^{LR}_{t\pm1}),\\
    \widetilde{h}^{\{f,b\}}_{t}&\!=\!\mathrm{warp}(h^{\{f,b\}}_{t\pm1},\, w^{\{f,b\}}_{t}),\\
    \widehat{h}^{\{f,b\}}_{t}&\!=\!R_{\{f,b\}}(I^{LR}_{t},\, \widetilde{h}^{\{f,b\}}_{t}),
\end{split}
\label{eq:resout}
\end{equation}
where $\mathrm{warp}(,)$ denotes warping operation, and $w^{\smash{\{f,b\}}}_{t}$ is the optical flow estimated by the flow estimation network $S$.
Note that the temporally aggregated features $\widehat{h}^{\smash{\{f,b\}}}_{t}$ contains details aggregated from multiple LR features, as well as temporal Ref features propagated from neighboring cells.

Now we propose the reference alignment and propagation module
for each cell $F_f$ and $F_b$ to fuse the current Ref frame $I^{\smash{Ref}}_{t}$ into temporally aggregated features $\widehat{h}^{\smash{\{f,b\}}}_{t}$.

\subsection{Reference Alignment and Propagation}
\label{sec:RAP}

\begin{figure}[t]
\begin{center}
\includegraphics [width=1.0\linewidth] {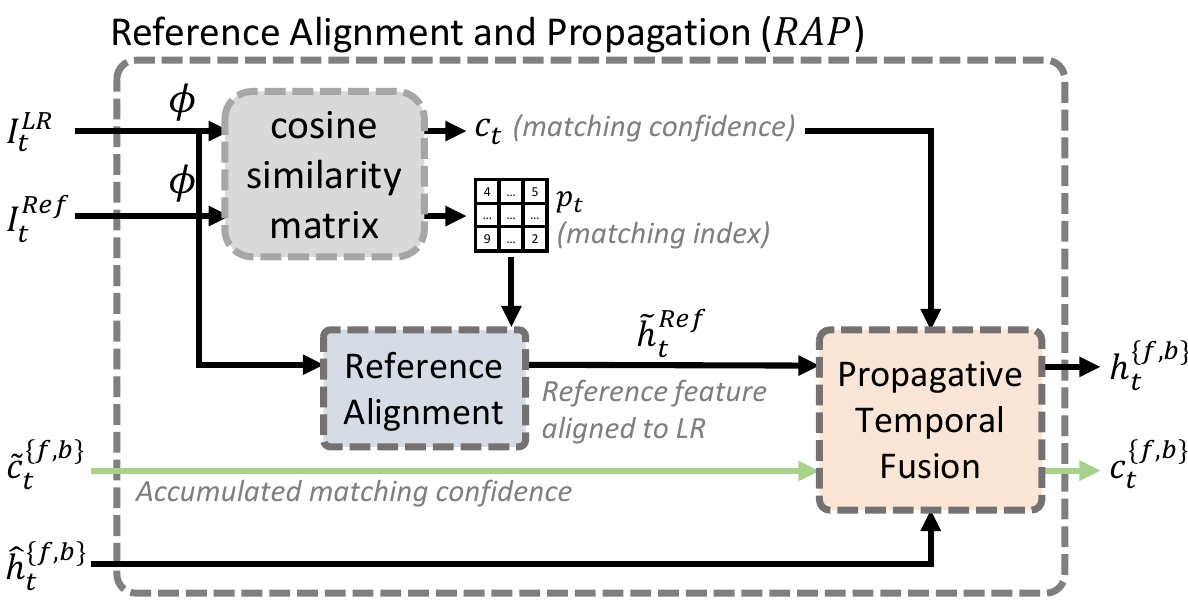}
\end{center}
\vspace{-0.6cm}
  \caption{The reference alignment and propagation module.}
\label{fig:RAP}
\vspace{-9pt}
\end{figure}

Our reference alignment and propagation module (\cref{fig:RAP}) consists of three sub-modules: cosine similarity, reference alignment, and propagative temporal fusion modules. %
The cosine similarity module computes a cosine similarity matrix between the Ref frame $I_{t}^{\smash{Ref}}$ and target LR frames $I_{t}^{\smash{LR}}$ and computes an index map $p_t$ and a confidence map $c_t$ needed for the other two sub-modules.
The reference alignment module extracts a feature map from the current Ref frame $I_t^{\smash{Ref}}$ and warps the feature map to $I_{t}^{\smash{LR}}$ using the index map $p_t$.
Then, the propagative temporal fusion module fuses the aligned Ref features with the \emph{temporally aggregated features} $\widehat{h}^{\smash{\{f,b\}}}_{t}$.
In the following, we describe each module in more detail.

\pagebreak
\paragraph{Cosine Similarity Module}
To compute an index map $p_t$ and a confidence map $c_t$,
we first embed $I^{\smash{LR}}_t$ and $I^{\smash{Ref}}_{t\downarrow}$ into the feature space by a shared encoder $\phi$~\cite{Simonyan15VGG}, where $\downarrow$ denotes the downsampling operator.
Then, we extract $\text{3}\!\times\!\text{3}$ patches from the LR and Ref feature maps with stride 1 and compute a cosine similarity matrix $C$ between them,
such that $C_{i,j}$ is a similarity between the $i$-th patch of the LR feature map and the $j$-th patch of the Ref feature map.
The matching index map $p$ and confidence map $c$ is then computed as:
\begin{equation}
    {p}_{t,i}\!=\!\argmax_j C_{i,j},\;\;\;\;\;\;\;\;{c}_{t,i} \!=\!\max_j C_{i,j},
\label{eq:matching}
\end{equation}
where $p_{t,i}$ is the patch index of Ref features $\phi(I^{\smash{Ref}}_{t\downarrow})$ that is the most relevant to the $i$-th patch of LR features $\phi(I^{\smash{LR}}_t)$, and $c_{t,i}$ is their matching confidence, respectively.

\vspace{-12pt}
\paragraph{Reference Alignment Module}
We use the reference alignment module proposed in~\cite{wang2021DCSR} to obtain Ref features aligned to $I^{\smash{LR}}_t$, which will be used for the fusion later.
The module first takes $I^{\smash{Ref}}_t$ and extracts Ref features $h^{\smash{Ref}}_t$.
Then, using the matching index map $p_t$ (\cref{eq:matching}), we warp patches of Ref features $h^{\smash{Ref}}_t$ to coarsely align the features to the current LR frame $I^{\smash{LR}}_t$~\cite{yang2020TTSR, Xie2020FeatureRM}.
Finally, the module compensates for possible inter-patch misalignment (\eg, scale and rotation) in the coarsely aligned Ref features using the patch-wise affine spatial transformer~\cite{Jaderberg2015STN, dai2017deformconv}.
We denote the final aligned Ref features as $\widetilde{h}^{\smash{Ref}}_t$.

\begin{figure}[t]
\begin{center}
\includegraphics [width=1.0\linewidth] {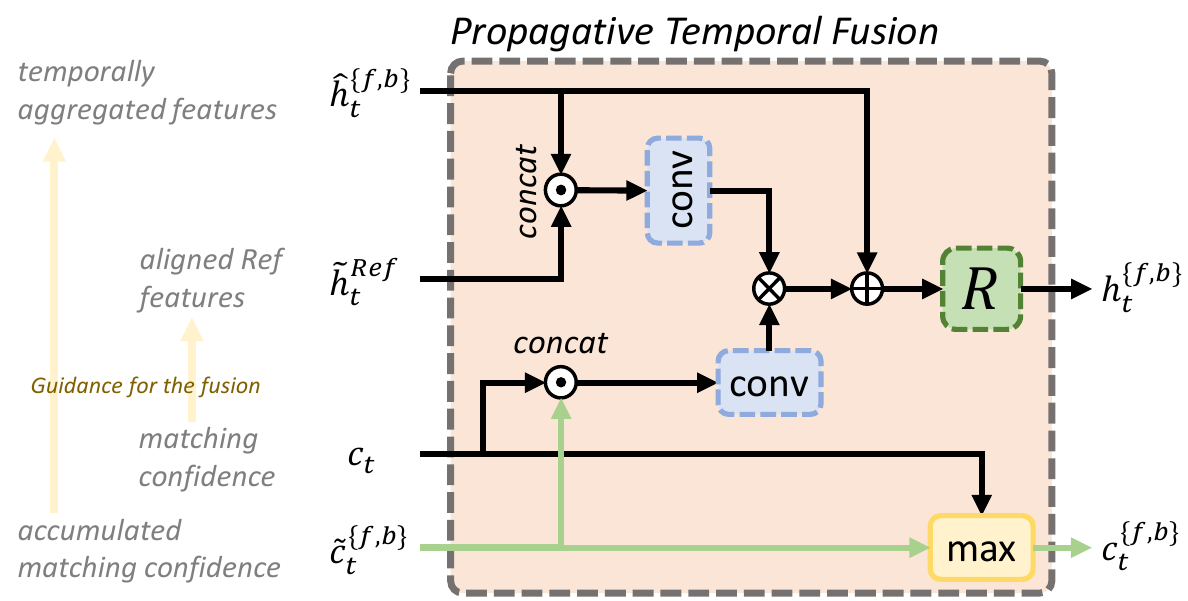}
\end{center}
\vspace{-0.6cm}
  \caption{The propagative temporal fusion module.}
\label{fig:PTF}
\vspace{-9pt}
\end{figure}

\vspace{-12pt}
\paragraph{Propagative Temporal Fusion Module}
Finally, we propose the propagative temporal fusion module that fuses the aligned Ref features $\widetilde{h}^{\smash{Ref}}_t$ with the \emph{temporally aggregated features} $\widehat{h}^{\smash{\{f,b\}}}_{t}$ and propagates the fused features $h^{\smash{\{f,b\}}}_{t}$ to the next cell (\cref{fig:PTF}).
Note that aligned Ref features $\widetilde{h}^{\smash{Ref}}_t$ contain Ref features at the current time step, while the temporally aggregated features $\widehat{h}^{\smash{\{f,b\}}}_{t}$ contain aggregated temporal Ref features propagated from neighboring recurrent cells.
For the successful fusion, the propagative temporal fusion module has to fuse $\widehat{h}^{\smash{\{f,b\}}}_{t}$ and $\widetilde{h}^{\smash{Ref}}_t$ in the way of selecting the Ref features better aligned to the target frame so that well-matched Ref features can keep propagating to the next cell.
Otherwise, erroneous Ref features can be accumulated in the pipeline, leading to blurry results.

However, a na\"ive fusion of the Ref features $\widetilde{h}^{\smash{Ref}}_t$ is error-prone, as matching is not necessarily accurate.
Inspired by\cite{yang2020TTSR, wang2021DCSR}, we thus perform feature fusion guided by the matching confidences $c_t$,
which guides the fusion module to select only well-matched features in $\widetilde{h}^{\smash{Ref}}_t$.
The fusion module also needs a guidance for propagated Ref features aggregated in $\widehat{h}^{\smash{\{f,b\}}}_{t}$.
The guidance should accommodate temporal information that coincides with propagated Ref features maintained in the propagation pipeline.
To this end, we accumulate matching confidences throughout the propagation pipeline and use the accumulated confidence as the guidance for the temporally aggregated features $\widehat{h}^{\smash{\{f,b\}}}_{t}$ during the fusion. Formally, we have:
\begin{equation}
\begin{split}
    \widetilde{c}^{\,\{f,b\}}_{t}&\!=\!\mathrm{warp}(c^{\{f,b\}}_{t\pm1},\, w^{\{f,b\}}_{t}),
\end{split}
\end{equation}
where $c^{\smash{\{f,b\}}}_{t\pm1}$ is the accumulated matching confidence propagated from neighboring cells. We align the confidence to obtain $\widetilde{c}^{\,\smash{\{f,b\}}}_{t}$ using the optical flow pre-computed in \cref{eq:resout}.

For the fusion, we provide matching confidence $c_t$ computed between the current target and reference frames, and we also provide aligned matching confidence $\widetilde{c}^{\,\smash{\{f,b\}}}_{t}$ propagated from neighboring recurrent cells as guidance.
The matching confidences are embedded with a convolution layer to consider matching scores of neighboring patches for providing more accurate guidance during the fusion~\cite{wang2021DCSR}.
Formally, the fusion process is defined as:
\begin{equation}
\begin{split}
    h^{\{f,b\}}_t\!=\!\{&\mathrm{conv}([c_{t},\,\widetilde{c}^{\,\{f,b\}}_{t}])\;\otimes \\ &\mathrm{conv}([\widetilde{h}^{Ref}_t,\,\widehat{h}^{\{f,b\}}_{t}])\} + \widehat{h}^{\{f,b\}}_{t},
\end{split}
\label{eq:PTF}
\end{equation}
where $[,]$ and $\otimes$ indicate concatenation operation and element-wise multiplication, respectively.

For the next cell, we use $\max(,)$ operation to accumulate $c_t$ into $\widetilde{c}^{\,\smash{\{f,b\}}}_{t}$ and pick up a larger confidence score.
The accumulation process is defined as:
\begin{equation}
    c^{\{f,b\}}_{t}\!=\!\max(c_t,\, \widetilde{c}^{\,\{f,b\}}_{t}).
\end{equation}
Max operation between $c_t$ and $\widetilde{c}^{\,\smash{\{f,b\}}}_{t}$ indirectly imposes the propagative temporal fusion module to selectively fuse and propagate the better matched features between corresponding features, $\widetilde{h}^{\smash{Ref}}_t$ and $\widehat{h}^{\smash{\{f,b\}}}_{t}$, respectively.

\section{Training Strategy for Real-World 4$\bm{\times}$VSR}
\label{sec:Real8K}

To train our network, we propose the RealMCVSR dataset, which consists of triplets of ultra-wide, wide-angle, and telephoto videos, where wide-angle and telephoto videos have the same size as ultra-wide videos, but their resolutions are 2$\times$ and 4$\times$ that of ultra-wide videos. The detail of the dataset is given in \cref{sec:experiments}.
Given video triplets, we train our network to perform 4$\times$ super-resolution of an ultra-wide HD video with a wide-angle video as a Ref video for obtaining an 8K video.
The resulting 8K video has the same resolution as a telephoto video, but 16$\times$ larger in size.

It is worth noting that we use only a wide-angle video as a Ref video.
While it may look reasonable to use a telephoto video as an additional Ref video to achieve the resolution of a telephoto video,
we found that it does not improve the super-resolution quality much because a telephoto video covers only 1/16 the area of an ultra-wide video.
A detailed discussion and experiments are provided in the supplement.

Training our network to produce 8K videos is not trivial as there are no ground-truth 8K videos.
While we have wide-angle and telephoto videos, they neither cover the entire area
nor perfectly align with an ultra-wide video.
To overcome this, we propose a novel training strategy that fully exploits wide-angle and telephoto videos.

Our training strategy consists of pre-training and adaptation stages.
In the pre-training stage, we downsample ultra-wide and wide-angle videos 4$\times$. %
We then train the network to 4$\times$ super-resolve a downsampled ultra-wide video using a downsampled wide-angle video as a reference.
The training is done in a supervised manner using the original ultra-wide video as the ground-truth.
Finally, in the adaptation stage, we fine-tune the network to adapt it to real-world videos of the original sizes.
This stage uses a telephoto video as supervision to train the network to recover high-frequency details of a telephoto video.
The following subsections describe each stage in more detail.

\subsection{Pre-training Stage}
\label{ssec:pretraining}

In this stage, we train our network using two loss functions: a reconstruction loss motivated by \cite{mechrez2018contextual, mechrez2018Learning, wang2021DCSR} and a multi-Ref fidelity loss.
The reconstruction loss minimizes the low- and high-frequency differences between a super-resolved ultra-wide frame $I^{SR}_t$ and the ground-truth ultra-wide frame $I^{HR}_t$. The reconstruction loss $\ell_{\mathit{rec}}$ is defined as:
\begingroup
\belowdisplayskip=\belowdisplayshortskip
\abovedisplayskip=\abovedisplayshortskip
\begin{equation}
    \ell_{\mathit{rec}}\!=\!\|I^{SR}_{t,blur}-I^{HR}_{t,blur}\|+\lambda_{\mathit{rec}}\sum_i\delta_i(I^{SR}_t,\, I^{HR}_t),
\label{eq:rec}
\end{equation}
\endgroup
where the subscript $blur$ indicates a filtering operation with a $\text{3}\!\times\!\text{3}$ Gaussian kernel with $\sigma\texttt{=}1.0$ and
$\lambda_{\mathit{rec}}$ is a weight for the second term.
$\delta_i(X, Y)\texttt{=}\min_j\mathbb{D}(x_i,y_j)$ is the contextual loss~\cite{mechrez2018contextual} that measures the distance between the pixel $x_i$ in $X$ and its most similar pixel $y_j$ in $Y$ under some feature distance measure $\mathbb{D}$, \eg, a perceptual distance~\cite{mechrez2018contextual, mechrez2018Learning, zhang2019zoom}.

In the first term on the right-hand side in \Eq{\ref{eq:rec}}, filtering frames with Gaussian kernels imposes results to follow low-frequency structures of a ground-truth ultra-wide frame $I^{HR}_t$.
The second term enforces the network to follow the high-frequency details of $I^{HR}_t$.
Note that in the second term, we use the contextual loss even for aligned pairs $I^{SR}_t$ and $I^{HR}_t$, as the loss is verified to be better in boosting the perceptual quality than the perceptual loss~\cite{Johnson2016percept} designed for aligned pairs~\cite{mechrez2018Learning}.

To guide the network to take advantage of multiple Ref frames, we encourage Ref features to keep propagating from one to the next cells.
Motivated by \cite{wang2021DCSR}, we propose a multi-Ref fidelity loss.
Given a super-resolved ultra-wide frame $I^{SR}_t$ and ground-truth wide-angle frames $I^{\smash{Ref_{HR}}}_{t\in \Omega}$, the multi-Ref fidelity loss is defined as:
\begin{equation}
\begin{split}
    \ell_{\mathit{Mfid}}\!=\!\frac{\sum_{t'\in \Omega}\sum_i\delta_i(I^{SR}_t,\, I^{Ref_{HR}}_{t'})\cdot c_{t'\!,i}}{\sum_{t'\in \Omega}\sum_i c_{t'\!,i}},
\end{split}
\label{eq:TFID}
\end{equation}
where $\Omega\texttt{=}[t\texttt{-}\frac{k-1}{2},\,\dots,\, t\texttt{+}\frac{k-1}{2}]$ is set of frame indices in a temporal window of size $k$. We use $k\texttt{=}7$ in practice.
Here, $c_{t'\!,i}$ is the matching confidence used for weighting the distance $\delta_i(I^{SR}_t,\,I^{\smash{Ref_{HR}}}_{t'})$.
Specifically, during training, pixels of $I^{SR}_t$ with higher matching confidence $c_{t'\!,i}$ are assigned with larger weights for optimization. 
\cref{eq:TFID} enables our network to effectively utilize multiple Ref frames $I^{\smash{Ref}}_{t\in \Omega}$ and keep the details of multiple Ref frames to flow through the propagation pipeline.
Our loss for the pre-training stage is defined as:
\begin{equation}
    \ell_{\mathit{pre}}\!=\!\ell_{\mathit{rec}}(I^{SR}_t,\, I^{HR}_t)+\lambda_{\mathit{pre}}\ell_{\mathit{Mfid}}(I^{SR}_t,\, I^{Ref_{HR}}_{t\in \Omega}).
\label{eq:pretraining}
\end{equation}
where $\lambda_{pre}$ is a weight for the multi-Ref fidelity loss for the pre-training stage.

\subsection{Adaptation Stage for Real-world 4$\bm{\times}$VSR}
\label{ssec:adaptation}

For adaptation, our network takes real-world ultra-wide $I^{UW}_t$ and wide-angle $I^{Wide}_t$ HD frames as LR and Ref frames, respectively.
As in the pre-training stage, the adaptation stage separately handles low- and high-frequency of a super-resolved ultra-wide frame $I^{SR}_t$.
However, as there is no ground-truth frame available for $I^{SR}_t$, we downsample $I^{SR}_t$ and use the input ultra-wide frame $I^{UW}_t$ as the supervision for recovering low-frequency structures.
For recovering high-frequency details, we directly utilize telephoto frames $I^{Tele}_{t\in \Omega}$ as the supervision for the proposed multi-Ref fidelity loss $\ell_{\mathit{Mfid}}$.
The adaptation loss is defined as:
\begin{equation}
    \ell_{\mathit{8K}}\!=\!||I^{SR}_{t\downarrow,blur}-I^{UW}_{t,blur}||+\lambda_{\mathit{8K}}\ell_{\mathit{Mfid}}(I^{SR}_t,\, I^{Tele}_{t\in \Omega}),
\label{eq:real8K}
\end{equation}
where $\lambda_{\mathit{8K}}$ is a weight for the multi-Ref fidelity loss for the adaptation stage.
The first term imposes our network to reconstruct low-frequency structures of input ultra-wide frames, and the second term trains our network to transfer the finest high-frequency details of telephoto frames.

\begin{table}[t]
\centering
\aboverulesep = 0mm %
\belowrulesep = 0mm %

\small
\setlength\tabcolsep{0pt}

\begin{tabular}{
>{\centering}p{0.2\columnwidth}
>{\centering}p{0.2\columnwidth}
>{\centering}p{0.2\columnwidth}
>{\centering}p{0.2\columnwidth}
>{\centering\arraybackslash}p{0.2\columnwidth}
}

$\ell_{\mathit{Mfid}}$ & PTF & PSNR$\uparrow$\! & SSIM$\uparrow$\! & Params (M)\\
\toprule
\rowcolor{lightlightgray}& & 30.71 & 0.894 & 4.2768 \\ %
\cm & & 31.31 & 0.913 & 4.2768 \\ 
\rowcolor{lightlightgray}\cm & \cm & \textbf{31.68} & \textbf{0.914} & 4.2772 \\ %

\bottomrule
\end{tabular}

\vspace{-0.25cm}
\caption{Quantitative ablation study. The first row corresponds to the baseline model. $\ell_{\mathit{Mfid}}$ and PTF indicate the models trained with \cref{eq:TFID} and propagative temporal fusion module, respectively.}
\label{tbl:ablation}
\vspace{-12pt}
\end{table}

\section{Experiments}
\label{sec:experiments}
\paragraph{RealMCVSR Dataset}
Our RealMCVSR dataset provides real-world HD video triplets concurrently recorded by Apple iPhone 12 Pro Max equipped with triple cameras having fixed focal lengths: ultra-wide (30mm), wide-angle (59mm), and telephoto (147mm). 
To concurrently record video triplets, we built an iOS app that provides full control over exposure parameters (\ie, shutter speed and ISO) of the cameras.
For recording each scene, we set the cameras in the auto-exposure mode, where the shutter speeds of the three cameras are synced to avoid varying motion blur across a video triplet. ISOs are adjusted accordingly for each camera to pick up the same exposure.
Each video is saved in the MOV format using HEVC/H.265 encoding with the HD resolution ($\text{1080}\!\times\!\text{1920}$).
The dataset contains triplets of 161 video clips with 23,107 frames in total.
The video triplets are split into training, validation, and testing sets, each of which has 137, 8, and 16 triplets with 19,426, 1,141, and 2,540 frames, respectively. %

\vspace{-12pt}
\paragraph{Implementation}
The network is trained using rectified-Adam~\cite{Liu2020RADAM} with an initial learning rate $2.0\!\times\!10^{\texttt{-}4}$, which is steadily decreased
to $1.0\!\times\!10^{\texttt{-}6}$ using the cosine annealing strategy~\cite{Ilya2017CA}.
The network is trained for 300k and 50k iterations for the pre-training and adaptation stages, respectively, with $\lambda_{\mathit{rec}}\,\texttt{=}\,0.01$, $\lambda_{pre}\,\texttt{=}\,0.05$, and $\lambda_{8K}\,\texttt{=}\,0.1$.
For each iteration, we randomly sample batches of frame triplets from the RealMCVSR training set. 
For the pre-training stage, we downsample ultra-wide LR and wide-angle Ref frames 4$\times$ using bicubic downsampling provided by MATLAB function $\mathrm{imresize}$.
We crop patches from each frame in a triplet to have overlapped contents and apply random translation on each crop window.
Then, ultra-wide LR frames are cropped to $\text{64}\!\times\!\text{64}$ and $\text{128}\!\times\!\text{128}$ patches for the pre-training and adaptation stages, respectively.
Wide-angle and telephoto Ref frames are cropped into patches of 2$\times$ and 4$\times$ the patch size of LR patches, respectively.

\begin{figure}[t]
\centering
    \def \ext {.pdf} %
    \setlength\tabcolsep{1pt}    
    \def \wb {0.25} %
    \def \ws {0.176} %

    \def \corlc {0.23} %
    \def \corbc {0.23} %
    \def \corrtc {0.54} %
    \def \corrtyc {0.54} %

    \def \img {0088/0141\ext}
    \def \imgw {0088/0141}
    \def \imgr {0088/0141_red\ext}
    \def \imgg {0088/0141_green\ext}
    \def \corl {0.555} %
    \def \corb {0.52} %
    \def \corrt {0.05} %
    \def \corrty {0.0764375} %
    
    \def \corls {0.346} %
    \def \corbs {0.78} %
    \def \corrts {0.04} %
    \def \corrtys {0.06115} %
    \footnotesize
    \begin{tabular}{ccccc}

    LR$\uparrow$ / Ref$\downarrow$ &
    Bicubic &
    Baseline &
    $\ell_\mathit{Mfid}$ &
    $\ell_\mathit{Mfid}$+PTF \\
    
    \toprule

    \begin{tikzpicture}
    \node[anchor=south west,inner sep=0] (image) at (0,0) {\includegraphics[width=\wb\columnwidth]{data/experiments/ablation/figure/Input/UW/\img}};
    \begin{scope}[x={(image.south east)},y={(image.north west)}]                                                 
        \draw[white, densely dashed, semithick] ([shift={(0.01, 0.01)}]\corlc, \corbc) rectangle ([shift={(-0.01, -0.01)}]{\corlc+\corrtc}, {\corbc+\corrtyc});
    \end{scope}
    \end{tikzpicture} 
    &
    \includegraphics[width=\ws\columnwidth,cfbox=red 0.8pt -0.8pt 0pt]{data/experiments/ablation/figure/Bicubic/\imgr}
    &
    \includegraphics[width=\ws\columnwidth,cfbox=red 0.8pt -0.8pt 0pt]{data/experiments/ablation/figure/FID/\imgr}
    &
    \includegraphics[width=\ws\columnwidth,cfbox=red 0.8pt -0.8pt 0pt]{data/experiments/ablation/figure/TFID/\imgr}
    &
    \includegraphics[width=\ws\columnwidth,cfbox=red 0.8pt -0.8pt 0pt]{data/experiments/ablation/figure/PTF/\imgr}
    
    \\[-0.01in]
    
    \begin{tikzpicture}
    \node[anchor=south west,inner sep=0] (image) at (0,0) {\includegraphics[trim=0 0 0 0, clip,  width=\wb\columnwidth]{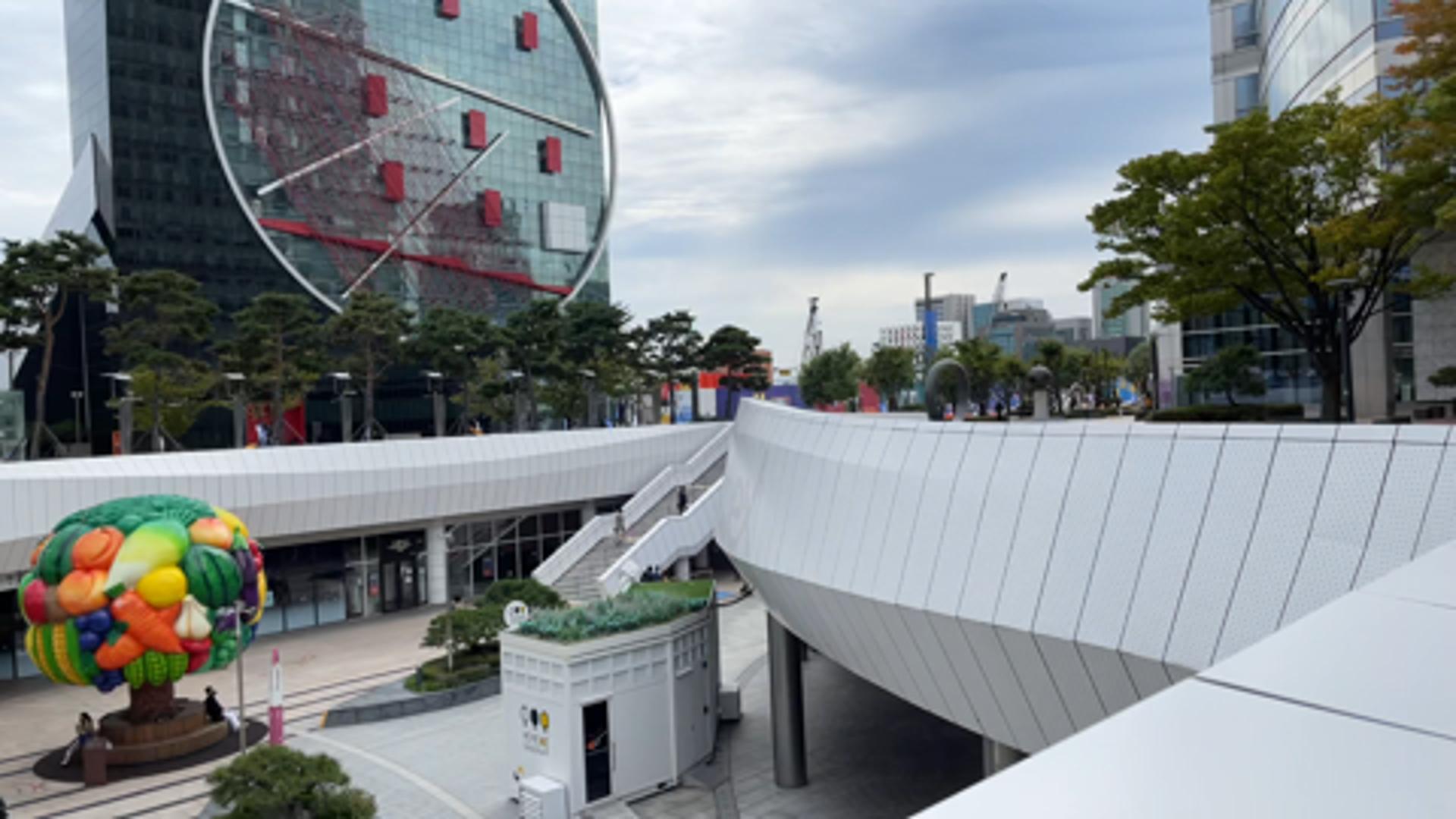}};
    \begin{scope}[x={(image.south east)},y={(image.north west)}]                                                 
        \draw[white, densely dashed, thick] ([shift={(0.007, 0.012)}]0, 0) rectangle ([shift={(-0.007, -0.012)}]{1.0}, {1.0});
    \end{scope}
    \end{tikzpicture}
    &
    
    \includegraphics[width=\ws\columnwidth,cfbox=green 0.8pt -0.8pt 0pt]{data/experiments/ablation/figure/Bicubic/\imgg}
    &
    \includegraphics[width=\ws\columnwidth,cfbox=green 0.8pt -0.8pt 0pt]{data/experiments/ablation/figure/FID/\imgg}
    &
    \includegraphics[width=\ws\columnwidth,cfbox=green 0.8pt -0.8pt 0pt]{data/experiments/ablation/figure/TFID/\imgg}
    &
    \includegraphics[width=\ws\columnwidth,cfbox=green 0.8pt -0.8pt 0pt]{data/experiments/ablation/figure/PTF/\imgg}
    
    \\[-0.03in]
    \midrule[0.26mm]
    
    \end{tabular}

    \def \corl {0.564} %
    \def \corb {0.38} %
    \def \corrt {0.04} %
    \def \corrty {0.061375} %
    
    \def \corls {0.945} %
    \def \corbs {0.863} %
    \def \corrts {0.047} %
    \def \corrtys {0.0736} %
    
    \def \corlc {0.23} %
    \def \corbc {0.23} %
    \def \corrtc {0.54} %
    \def \corrtyc {0.54} %
    
    \def \corlcp {0.23} %
    \def \corbcp {0.0} %
    \def \corrtcp {0.54} %
    \def \corrtycp {1} %
    
    \def \img {0010/0124\ext}
    \def \imgr {0010/0124_red\ext}
    \def \imgg {0010/0124_green\ext}
    
    \footnotesize
    \begin{tabular}{ccccc}
    \\[-0.149in]
    
    \begin{tikzpicture}
    \node[anchor=south west,inner sep=0] (image) at (0,0) {\includegraphics[width=\wb\columnwidth]{data/experiments/ablation/figure/Input/UW/\img}};
    \begin{scope}[x={(image.south east)},y={(image.north west)}]                                                 
        \draw[white, densely dashed, semithick] ([shift={(0.01, 0.01)}]\corlc, \corbc) rectangle ([shift={(-0.01, -0.01)}]{\corlc+\corrtc}, {\corbc+\corrtyc});
        \begin{scope}[x={(image.south east)},y={(image.north west)}]
            \coordinate (A) at (\corlc+0.01,0.01);
            \coordinate (B) at (\corlc+0.01,\corbc-0.01);
            \coordinate (C) at (\corlc+0.01,\corbc+\corrtyc+0.01);
            \coordinate (D) at (\corlc+0.01,1-0.01);
            \coordinate (E) at (\corlc+\corrtc-0.01,0.01);
            \coordinate (F) at (\corlc+\corrtc-0.01,\corbc-0.01);
            \coordinate (G) at (\corlc+\corrtc-0.01,\corbc+\corrtyc+0.01);
            \coordinate (H) at (\corlc+\corrtc-0.01,1-0.01);
            \draw[white,densely dashed,line width=0.005mm] (A) -- (B);
            \draw[white,densely dashed,line width=0.005mm] (C) -- (D);
            \draw[white,densely dashed,line width=0.005mm] (E) -- (F);
            \draw[white,densely dashed,line width=0.005mm] (G) -- (H);
        \end{scope}
    
    \end{scope}
    \end{tikzpicture} 
    &
    \includegraphics[width=\ws\columnwidth,cfbox=red 0.8pt -0.8pt 0pt]{data/experiments/ablation/figure/Bicubic/\imgr}
    &
    \includegraphics[width=\ws\columnwidth,cfbox=red 0.8pt -0.8pt 0pt]{data/experiments/ablation/figure/FID/\imgr}
    &
    \includegraphics[width=\ws\columnwidth,cfbox=red 0.8pt -0.8pt 0pt]{data/experiments/ablation/figure/TFID/\imgr}
    &
    \includegraphics[width=\ws\columnwidth,cfbox=red 0.8pt -0.8pt 0pt]{data/experiments/ablation/figure/PTF/\imgr}
    
    \\[-0.01in]
    
    \begin{tikzpicture}
    \node[anchor=south west,inner sep=0] (image) at (0,0)    {\includegraphics[trim=0 0 0 0, clip,  width=\wb\columnwidth]{data/experiments/ablation/figure/Input/W/\img}};
    \begin{scope}[x={(image.south east)},y={(image.north west)}]                                                 
        \draw[white, densely dashed, thick] ([shift={(0.007, 0.012)}]0, 0) rectangle ([shift={(-0.007, -0.012)}]{1.0}, {1.0});
    \end{scope}
    \end{tikzpicture}
    &
    
    \includegraphics[width=\ws\columnwidth,cfbox=green 0.8pt -0.8pt 0pt]{data/experiments/ablation/figure/Bicubic/\imgg}
    &
    \includegraphics[width=\ws\columnwidth,cfbox=green 0.8pt -0.8pt 0pt]{data/experiments/ablation/figure/FID/\imgg}
    &
    \includegraphics[width=\ws\columnwidth,cfbox=green 0.8pt -0.8pt 0pt]{data/experiments/ablation/figure/TFID/\imgg}
    &
    \includegraphics[width=\ws\columnwidth,cfbox=green 0.8pt -0.8pt 0pt]{data/experiments/ablation/figure/PTF/\imgg}

    \end{tabular}
\vspace{-0.4cm}
\caption{Qualitative ablation study.
The first column shows LR and Ref real-world HD inputs.
For the rest of the columns, we show zoomed-in cropped 4$\times$SR results of different combinations of modules (\Tbl{\ref{tbl:ablation}}).
Red and green boxes indicate inside and outside the overlapped FoV between LR and Ref frames, respectively. }
\label{fig:ablation}
\vspace{-12pt}
\end{figure}

\subsection{Ablation Study}
To analyze the effect of each component of our model, we conduct ablation studies.
First, we validate the effects of the propagative temporal fusion module (\cref{eq:PTF}) and multi-Ref fidelity loss $\ell_{\mathit{Mfid}}$ (\cref{eq:TFID}).
To this end, we compare the stripped-out baseline model with its two variants.
The baseline model is trained with $\ell_{\mathit{rec}}$ and $\ell_{\mathit{Mfid}}$, but we set the temporal window size $k\texttt{=}1$ for $\ell_{\mathit{Mfid}}$,
indicating only a single ground-truth Ref frame is used for computing the loss.
Regarding the propagative temporal fusion module, we use a modified one for the baseline model.
Specifically, \cref{eq:PTF} becomes:
\begin{equation}
\begin{split}
    h^{\{f,b\}}_t=\{\mathrm{conv}(c_{t})\otimes \mathrm{conv}([\widetilde{h}^{Ref}_t,\,\widehat{h}^{\{f,b\}}_{t}])\} + \widehat{h}^{\{f,b\}}_{t}.\nonumber
\end{split}
\end{equation}
For the other variants, we recover the key components one by one from the baseline model.
For the variant with $\ell_{\mathit{Mfid}}$, we train the baseline model with $\ell_{\mathit{rec}}$ and $\ell_{\mathit{Mfid}}$ with window size $k\texttt{=}7$.
For the last variant, we attach the propagative temporal fusion module.
For quantitative and qualitative comparison, we compare pre-trained models (\cref{ssec:pretraining}) and their fine-tuned models (\cref{ssec:adaptation}) on the proposed RealMCVSR test set, respectively.

\cref{tbl:ablation} shows quantitative results.
The table indicates that compared to the baseline model (the first row in the table), the model trained with $\ell_{\mathit{Mfid}}$ (the second row) shows much better VSR performance.
The model additionally equipped with the propagative temporal fusion module (the third row) achieves the best results in every measure.

\cref{fig:ablation} shows a qualitative comparison.
As shown in the figure,
the model trained with $\ell_{\mathit{Mfid}}$ (the fourth column of the figure) enhances
details inside (red box) and outside (green box) the overlapped FoV much better compared to the results of the baseline model (the third column).
The result confirms that $\ell_{\mathit{Mfid}}$ enforces temporal Ref features to keep streaming through the propagation pipeline to be utilized in reconstructing high-fidelity results.
The model attached with the propagative temporal fusion module shows accurately recovered structures and enhanced details for both inside and outside the overlapped FoV (the last column).
This demonstrates the propagative temporal fusion module promotes well-matched Ref features to be fused and to flow through the propagation pipeline.

\begin{figure}[t]
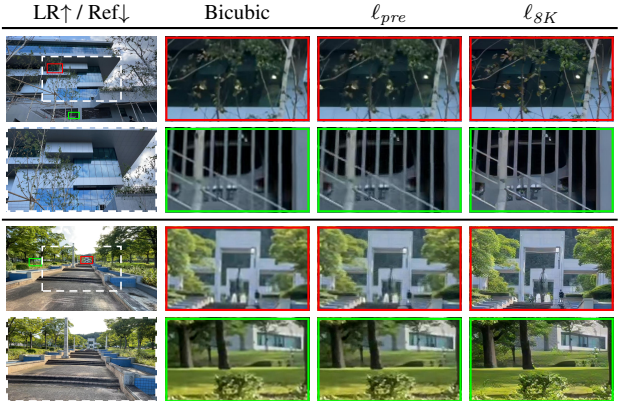

\centering
    \def \ext {.pdf} %
    \setlength\tabcolsep{1.5 pt}    
    \def \wb {0.24} %
    \def \ws {0.23} %

    \def \corlc {0.23} %
    \def \corbc {0.23} %
    \def \corrtc {0.54} %
    \def \corrtyc {0.54} %

    \def \img {0147/0071\ext}
    \def \imgw {0147/0071\ext}
    \def \imgr {0147/0071_red\ext}
    \def \imgg {0147/0071_green\ext}
    
    \def \imgt {0141/0014\ext}
    \def \imgwt {0141/0014\ext}
    \def \imgrt {0141/0014_red\ext}
    \def \imggt {0141/0014_green\ext}
    \def \corl {0.555} %
    \def \corb {0.52} %
    \def \corrt {0.05} %
    \def \corrty {0.0764375} %
    
    \def \corls {0.346} %
    \def \corbs {0.78} %
    \def \corrts {0.04} %
    \def \corrtys {0.06115} %
    \footnotesize
    \begin{tabular}{cccc}

    LR$\uparrow$ / Ref$\downarrow$ &
    Bicubic &
    $\ell_\mathit{pre}$ &
    $\ell_\mathit{8K}$ \\
    
    \toprule

    \begin{tikzpicture}
    \node[anchor=south west,inner sep=0] (image) at (0,0) {\includegraphics[width=\wb\columnwidth]{data/experiments/ablation/figure/Input/UW/\img}};
    \begin{scope}[x={(image.south east)},y={(image.north west)}]                                                 
        \draw[white, densely dashed, semithick] ([shift={(0.01, 0.01)}]\corlc, \corbc) rectangle ([shift={(-0.01, -0.01)}]{\corlc+\corrtc}, {\corbc+\corrtyc});
    \end{scope}
    \end{tikzpicture} 
    &
    \includegraphics[width=\ws\columnwidth,cfbox=red 0.8pt -0.8pt 0pt]{data/experiments/ablation/figure/Bicubic/\imgr}
    &
    \includegraphics[width=\ws\columnwidth,cfbox=red 0.8pt -0.8pt 0pt]{data/experiments/ablation/figure/L_rec/\imgr}
    &
    \includegraphics[width=\ws\columnwidth,cfbox=red 0.8pt -0.8pt 0pt]{data/experiments/ablation/figure/L_8K/\imgr}
    
    \\[-0.01in]
    
    \begin{tikzpicture}
    \node[anchor=south west,inner sep=0] (image) at (0,0) {\includegraphics[trim=0 0 0 0, clip,  width=\wb\columnwidth]{data/experiments/ablation/figure/Input/W/\imgw}};
    \begin{scope}[x={(image.south east)},y={(image.north west)}]                                                 
        \draw[white, densely dashed, thick] ([shift={(0.007, 0.012)}]0, 0) rectangle ([shift={(-0.007, -0.012)}]{1.0}, {1.0});
    \end{scope}
    \end{tikzpicture}
    &
    
    \includegraphics[width=\ws\columnwidth,cfbox=green 0.8pt -0.8pt 0pt]{data/experiments/ablation/figure/Bicubic/\imgg}
    &
    \includegraphics[width=\ws\columnwidth,cfbox=green 0.8pt -0.8pt 0pt]{data/experiments/ablation/figure/L_rec/\imgg}
    &
    \includegraphics[width=\ws\columnwidth,cfbox=green 0.8pt -0.8pt 0pt]{data/experiments/ablation/figure/L_8K/\imgg}
    
    \\[-0.03in]

    \midrule[0.26mm]
    \\[-0.14in]

    \begin{tikzpicture}
    \node[anchor=south west,inner sep=0] (image) at (0,0) {\includegraphics[width=\wb\columnwidth]{data/experiments/ablation/figure/Input/UW/\imgt}};
    \begin{scope}[x={(image.south east)},y={(image.north west)}]                                                 
        \draw[white, densely dashed, semithick] ([shift={(0.01, 0.01)}]\corlc, \corbc) rectangle ([shift={(-0.01, -0.01)}]{\corlc+\corrtc}, {\corbc+\corrtyc});
    \end{scope}
    \end{tikzpicture} 
    &
    \includegraphics[width=\ws\columnwidth,cfbox=red 0.8pt -0.8pt 0pt]{data/experiments/ablation/figure/Bicubic/\imgrt}
    &
    \includegraphics[width=\ws\columnwidth,cfbox=red 0.8pt -0.8pt 0pt]{data/experiments/ablation/figure/L_rec/\imgrt}
    &
    \includegraphics[width=\ws\columnwidth,cfbox=red 0.8pt -0.8pt 0pt]{data/experiments/ablation/figure/L_8K/\imgrt}
    
    \\[-0.01in]
    
    \begin{tikzpicture}
    \node[anchor=south west,inner sep=0] (image) at (0,0) {\includegraphics[trim=0 0 0 0, clip,  width=\wb\columnwidth]{data/experiments/ablation/figure/Input/W/\imgwt}};
    \begin{scope}[x={(image.south east)},y={(image.north west)}]                                                 
        \draw[white, densely dashed, thick] ([shift={(0.007, 0.012)}]0, 0) rectangle ([shift={(-0.007, -0.012)}]{1.0}, {1.0});
    \end{scope}
    \end{tikzpicture}
    &
    
    \includegraphics[width=\ws\columnwidth,cfbox=green 0.8pt -0.8pt 0pt]{data/experiments/ablation/figure/Bicubic/\imggt}
    &
    \includegraphics[width=\ws\columnwidth,cfbox=green 0.8pt -0.8pt 0pt]{data/experiments/ablation/figure/L_rec/\imggt}
    &
    \includegraphics[width=\ws\columnwidth,cfbox=green 0.8pt -0.8pt 0pt]{data/experiments/ablation/figure/L_8K/\imggt}

    \end{tabular}
\vspace{-0.4cm}
\caption{Ablation study on the two-stage training strategy.}
\label{fig:ablation_8K}
\vspace{-14pt}
\end{figure}

\pagebreak
We also validate the effects of the proposed training strategy.
Specifically, we qualitatively compare between the model pre-trained with the pre-training loss $\ell_{\mathit{pre}}$ (\cref{eq:pretraining}) and the model fine-tuned with the adaptation loss $\ell_{\mathit{8K}}$ (\cref{eq:real8K}).
For comparison, we show 8K VSR results given real-world HD videos.
Note that in the real-world scenario, there is no ground-truth available for a quantitative comparison.
\cref{fig:ablation_8K} shows the results.
The pre-trained model does not improve details of a real-world input, due to the domain gap between real-world inputs and downsampled inputs (the third column).
However, the fine-tuned model shows much higher fidelity results compared to the pre-trained model (the last column), thanks to the adaptation stage that trains the network to well adapt to real-world videos.

\subsection{Comparison on RealMCVSR Dataset}
In this section, we compare our method with previous state-of-the-art approaches: SRCNN~\cite{Dong2016SRCNN}, RCAN~\cite{zhang2018rcan}, TTSR~\cite{yang2020TTSR}, DCSR~\cite{wang2021DCSR}, EDVR~\cite{wang2019edvr}, BasicVSR~\cite{chan2021basicvsr}, and IconVSR~\cite{chan2021basicvsr}.
SRCNN and RCAN are SISR models that take only a single LR frame.
TTSR and DCSR are RefSR models fed with a pair of LR and Ref frames.
EDVR is a sliding window-based VSR model that takes multiple frames in a local temporal window.
BasicVSR and IconVSR are VSR models with a bidirectional recurrent framework, where each video frame is fed to a recurrent cell for each time step.
We train each model with the proposed RealMCVSR dataset and the code provided by the authors.

\vspace{-12pt}
\paragraph{Quantitative Comparison} 
\cref{tbl:comparison:4XHD} shows a quantitative comparison, where ultra-wide HD frames and their 4$\times$ downsampled ones are used as ground-truths and inputs, respectively.
For comparison, we use our model pre-trained with \cref{eq:pretraining} (Ours).
Moreover, to consider the trade-off between the model size and SR quality, we show the results of a smaller model with fewer parameters (Ours-\textit{small}) and a larger model (Ours-IR) attached with information-refill and coupled propagation modules proposed in \cite{chan2021basicvsr}.
We also compare our models trained only with the $\ell_1$ loss function (models indicated with -$\ell_1$), for a fair comparison with the previous models trained with pixel-based losses, such as $\ell_{1}$, $\ell_{2}$, and $\ell_{ch}$ (Charbonnier loss~\cite{lai2017LapSRN}),
which are known for having an advantage in PSNR over perceptual-based loss~\cite{Johnson2016percept}.

In \cref{tbl:comparison:4XHD}, while RefSR methods show a better performance than SISR methods, our methods outperform all previous ones.
Interestingly, VSR methods outperform RefSR methods that are additionally fed with Ref frames.
However, this is not particularly true if we measure the performance on the regions of the SR frame corresponding to different FoV ranges.
\cref{tbl:comparison:FOV} shows the results.
For comparison, we measure the SR quality for the region inside the overlapped FoV (0$\pct$--50$\pct$) between an ultra-wide SR and a wide-angle Ref frames.
For outside the overlapped FoV, we measure SR performance for the banded regions at different FoV ranges from the overlapped FoV (50$\pct$) to full FoV (100$\pct$).
In the table, DCSR~\cite{wang2021DCSR} outperforms IconVSR~\cite{chan2021basicvsr} for the overlapped FoV (0$\pct$--50$\pct$) between an input and Ref frames, while IconVSR outruns DCSR for the rest of the regions.
Our models exceed all models for all regions.

Note that in \cref{tbl:comparison:FOV}, our models show a performance gap between regions inside (0$\pct$--50$\pct$) and outside (50$\pct$--100$\pct$) the overlapped FoV.
However, compared to the PSNR/SSIM gap of DCSR (8.5$\pct$\,/\,4.2$\pct$), our models show much smaller gap (Ours-$\ell_1$\,:\,4.2$\pct$\,/\,1.8$\pct$ and Ours-IR-$\ell_{1}$\,:\,4.2$\pct$\,/\,1.6$\pct$). The result implies the proposed architecture effectively utilizes neighboring Ref features for recovering regions both inside and outside of the overlapped FoV.

\begin{table}[t]
\centering
\aboverulesep = 0.01mm %
\belowrulesep = 0.01mm %
\small
\setlength\tabcolsep{0pt}
\begin{tabularx}{\columnwidth}{@{\hspace{5pt}}l@{\hspace{7pt}}lYcYcYc}
& Model && PSNR$\uparrow$\! && SSIM$\uparrow$\! && Params (M) \\
\toprule
\multirow{3}{*}{\rotatebox[origin=c]{90}{\textbf{SISR}}} 
& Bicubic && 26.65 && 0.800 && - \\
& SRGAN~\cite{Ledig2017SRGAN} && 29.38 && 0.877 && 0.734 \\
&\cellcolor{lightlightgray}RCAN-{$\ell_{1}$}~\cite{zhang2018rcan} &\cellcolor{lightlightgray}& \cellcolor{lightlightgray}31.07 &\cellcolor{lightlightgray}& \cellcolor{lightlightgray}0.915 &\cellcolor{lightlightgray}& \cellcolor{lightlightgray}15.89 \\
\midrule[0.2pt]

\multirow{4}{*}{\rotatebox[origin=c]{90}{\textbf{RefSR}}} 
& TTSR~\cite{yang2020TTSR} && 30.31 && 0.905 && 6.730 \\
& TTSR-{$\ell_{1}$}~\cite{yang2020TTSR} && 30.83 && 0.911 && 6.730 \\
& DCSR~\cite{wang2021DCSR} && 30.63 && 0.895 && 5.419 \\
& \cellcolor{lightlightgray}DCSR-{$\ell_{1}$}~\cite{wang2021DCSR} &\cellcolor{lightlightgray}& \cellcolor{lightlightgray}32.43 &\cellcolor{lightlightgray}& \cellcolor{lightlightgray}0.933 &\cellcolor{lightlightgray}& \cellcolor{lightlightgray}5.419 \\
\midrule[0.2pt]

\multirow{4}{*}{\rotatebox[origin=c]{90}{\textbf{VSR}}} 
& EDVR-M-{$\ell_{ch}$}~\cite{wang2019edvr} && 33.26 && 0.946 && 3.317 \\
& EDVR-{$\ell_{ch}$}~\cite{wang2019edvr} && 33.47 && 0.948 && 20.63 \\
& BasicVSR-{$\ell_{ch}$}~\cite{chan2021basicvsr} && 33.66 && 0.951 && 4.851 \\
&\cellcolor{lightlightgray}IconVSR-{$\ell_{ch}$}~\cite{chan2021basicvsr}&\cellcolor{lightlightgray}&\cellcolor{lightlightgray}33.80&\cellcolor{lightlightgray}&\cellcolor{lightlightgray}0.951&\cellcolor{lightlightgray}&\cellcolor{lightlightgray}7.255\\
\midrule[0.2pt]

\multirow{6}{*}[-0.15\dimexpr 10\cmidrulewidth]{\rotatebox[origin=c]{90}{\textbf{RefVSR}}} 

& Ours-$\textit{small}$ && 31.63 && 0.912 && 1.052 \\ %
&\cellcolor{lightlightgray}Ours-$\textit{small}$-$\ell_{1}$ &\cellcolor{lightlightgray}&\cellcolor{lightlightgray}33.88 &\cellcolor{lightlightgray}&\cellcolor{lightlightgray}0.951 &\cellcolor{lightlightgray}&\cellcolor{lightlightgray}1.052 \\

\arrayrulecolor{gray}\cmidrule{2-8}\arrayrulecolor{black}
& Ours && 31.68 && 0.914 && 4.277 \\ %
& \cellcolor{lightlightgray}Ours-{$\ell_{1}$} &\cellcolor{lightlightgray}& \cellcolor{lightlightgray}34.74%
\cellcolor{lightlightgray}&\cellcolor{lightlightgray}& \cellcolor{lightlightgray}0.958 
\cellcolor{lightlightgray}&\cellcolor{lightlightgray}& \cellcolor{lightlightgray}4.277 \\

\arrayrulecolor{gray}\cmidrule{2-8}\arrayrulecolor{black}
& Ours-IR && 31.73 && 0.916 && 4.774 \\ %
&\cellcolor{lightlightgray}Ours-IR-{$\ell_{1}$} &\cellcolor{lightlightgray}&\cellcolor{lightlightgray}\textbf{34.86}
&\cellcolor{lightlightgray}&\cellcolor{lightlightgray}\textbf{0.959}
&\cellcolor{lightlightgray}&\cellcolor{lightlightgray}4.774 \\

\bottomrule
\end{tabularx}

\vspace{-0.25cm}
\caption{Quantitative evaluation on the RealMCVSR test set.}
\vspace{-14pt}
\label{tbl:comparison:4XHD}
\end{table}

\begin{table*}[t]
\centering
\aboverulesep = 0.11mm %
\belowrulesep = 0.1mm %

\setlength\tabcolsep{0pt}
\small
\begin{tabularx}{\textwidth}{@{\hspace{5pt}}l@{\hspace{10pt}}lYYcYcYcYcYcYcYcY}
\toprule
&\multirow{2}{*}[-0.03\dimexpr 30\cmidrulewidth]{Model}&&
\multicolumn{12}{c}{PSNR / SSIM measured for regions in the indicated FoV range} && \multirow{2}{*}[-0.03\dimexpr 30\cmidrulewidth]{\makecell{Params\\(M)}}& \\
&  &&& 0$\pct$--50$\pct$ && 50$\pct$--60$\pct$ && 50$\pct$--70$\pct$ && 50$\pct$--80$\pct$ && 50$\pct$--90$\pct$ && 50$\pct$--100$\pct$ &&&\\
\midrule[0.5pt]
\multirow{2}{*}[-0.03\dimexpr 30\cmidrulewidth]{\textbf{SISR}} & Bicubic &&& 25.38 / 0.757 && 26.30 / 0.785 && 26.42 / 0.789 && 26.71 / 0.798 && 26.99 / 0.801 && 27.29 / 0.815 && - &\\
& RCAN-$\ell_{1}$~\cite{zhang2018rcan} &&& 29.77 / 0.895 && 30.69 / 0.908 && 30.86 / 0.910 && 31.17 / 0.914 && 31.50 / 0.918 && 31.80 / 0.921 && 15.89 & \\

\midrule[0.1pt]
\multirow{1}{*}[-0.09\dimexpr 30\cmidrulewidth]{\textbf{RefSR}}
&\cellcolor{lightlightgray}DCSR-$\ell_{1}$~\cite{wang2021DCSR}
&\cellcolor{lightlightgray}&\cellcolor{lightlightgray}
&\cellcolor{lightlightgray}34.90 / 0.963
\cellcolor{lightlightgray}&\cellcolor{lightlightgray}&\cellcolor{lightlightgray}31.96 / 0.927
\cellcolor{lightlightgray}&\cellcolor{lightlightgray}\cellcolor{lightlightgray}&31.61 / 0.921
\cellcolor{lightlightgray}&\cellcolor{lightlightgray}&\cellcolor{lightlightgray}31.58 / 0.919
\cellcolor{lightlightgray}&\cellcolor{lightlightgray}&\cellcolor{lightlightgray}31.81 / 0.921
\cellcolor{lightlightgray}&\cellcolor{lightlightgray}&\cellcolor{lightlightgray}31.93 / 0.923
&\cellcolor{lightlightgray}&\cellcolor{lightlightgray}5.419
&\cellcolor{lightlightgray}\\

\midrule[0.1pt]
\multirow{1}{*}[-0.09\dimexpr 30\cmidrulewidth]{\textbf{VSR}} & IconVSR-$\ell_{ch}$~\cite{chan2021basicvsr}
&&& 32.79 / 0.946 && 33.43 / 0.949 && 33.60 / 0.950 && 33.89 / 0.951 && 34.19 / 0.953 && 34.40 / 0.953 && 7.255 &\\

\midrule[0.1pt]

\multirow{2}{*}[-0.03\dimexpr 30\cmidrulewidth]{\textbf{RefVSR}} 

&\cellcolor{lightlightgray}Ours-$\ell_{1}$
&\cellcolor{lightlightgray}&\cellcolor{lightlightgray}
&\cellcolor{lightlightgray}36.02 / 0.971
&\cellcolor{lightlightgray}&\cellcolor{lightlightgray}34.59 / 0.958
&\cellcolor{lightlightgray}&\cellcolor{lightlightgray}34.31 / 0.956
&\cellcolor{lightlightgray}&\cellcolor{lightlightgray}34.23 / 0.954
&\cellcolor{lightlightgray}&\cellcolor{lightlightgray}34.40 / 0.955
&\cellcolor{lightlightgray}&\cellcolor{lightlightgray}34.50 / 0.954
&\cellcolor{lightlightgray}&\cellcolor{lightlightgray}4.277
&\cellcolor{lightlightgray}\\

&\cellcolor{lightlightgray}Ours-IR-$\ell_{1}$
&\cellcolor{lightlightgray}&\cellcolor{lightlightgray}
&\cellcolor{lightlightgray}\textbf{36.14 / 0.971}
&\cellcolor{lightlightgray}&\cellcolor{lightlightgray}\textbf{34.66 / 0.959}
&\cellcolor{lightlightgray}&\cellcolor{lightlightgray}\textbf{34.40 / 0.956}
&\cellcolor{lightlightgray}&\cellcolor{lightlightgray}\textbf{34.34 / 0.955}
&\cellcolor{lightlightgray}&\cellcolor{lightlightgray}\textbf{34.52 / 0.955}
&\cellcolor{lightlightgray}&\cellcolor{lightlightgray}\textbf{34.63 / 0.955}
&\cellcolor{lightlightgray}&\cellcolor{lightlightgray}4.774
&\cellcolor{lightlightgray}\\

\bottomrule
\end{tabularx}

\vspace{-0.25cm}
\caption{Quantitative results measured with varying FoV range.
The center 50$\pct$ of FoV in an ultra-wide SR frame is overlapped with the FoV of a wide-angle reference frame.
Here, 0$\pct$--50$\pct$ indicates the region inside the overlapped FoV, and 50$\pct$--100$\pct$ is the region outside the overlapped FoV.
50$\pct$--60$\pct$ means the banded region between the center 50$\pct$ and 
60$\pct$ of an ultra-wide SR frame.
}
\label{tbl:comparison:FOV}
\vspace{-4pt}
\end{table*}

\begin{figure*}[t]
\centering
    \def \ext {.jpg} %
    \setlength\tabcolsep{1.5pt}
    
    \def \wb {0.20} %
    \def \ws {0.1509} %

    \def \corlc {0.23} %
    \def \corbc {0.23} %
    \def \corrtc {0.54} %
    \def \corrtyc {0.54} %

    \def \img {0121/0093}
    \def \imgr {0121/0093_red}
    \def \imgg {0121/0093_green}
    
    \footnotesize
    \begin{tabular}{cccccc}
    LR$\uparrow$ / Ref$\downarrow$ &
    (a) Bicubic &
    (c) RCAN \cite{zhang2018rcan} &
    (d) DCSR \cite{wang2021DCSR} &
    (e) IconVSR \cite{chan2021basicvsr} &
    (f) Ours \\
    \toprule
    \begin{tikzpicture}
    \node[anchor=south west,inner sep=0] (image) at (0,0) {\includegraphics[width=\wb\textwidth]{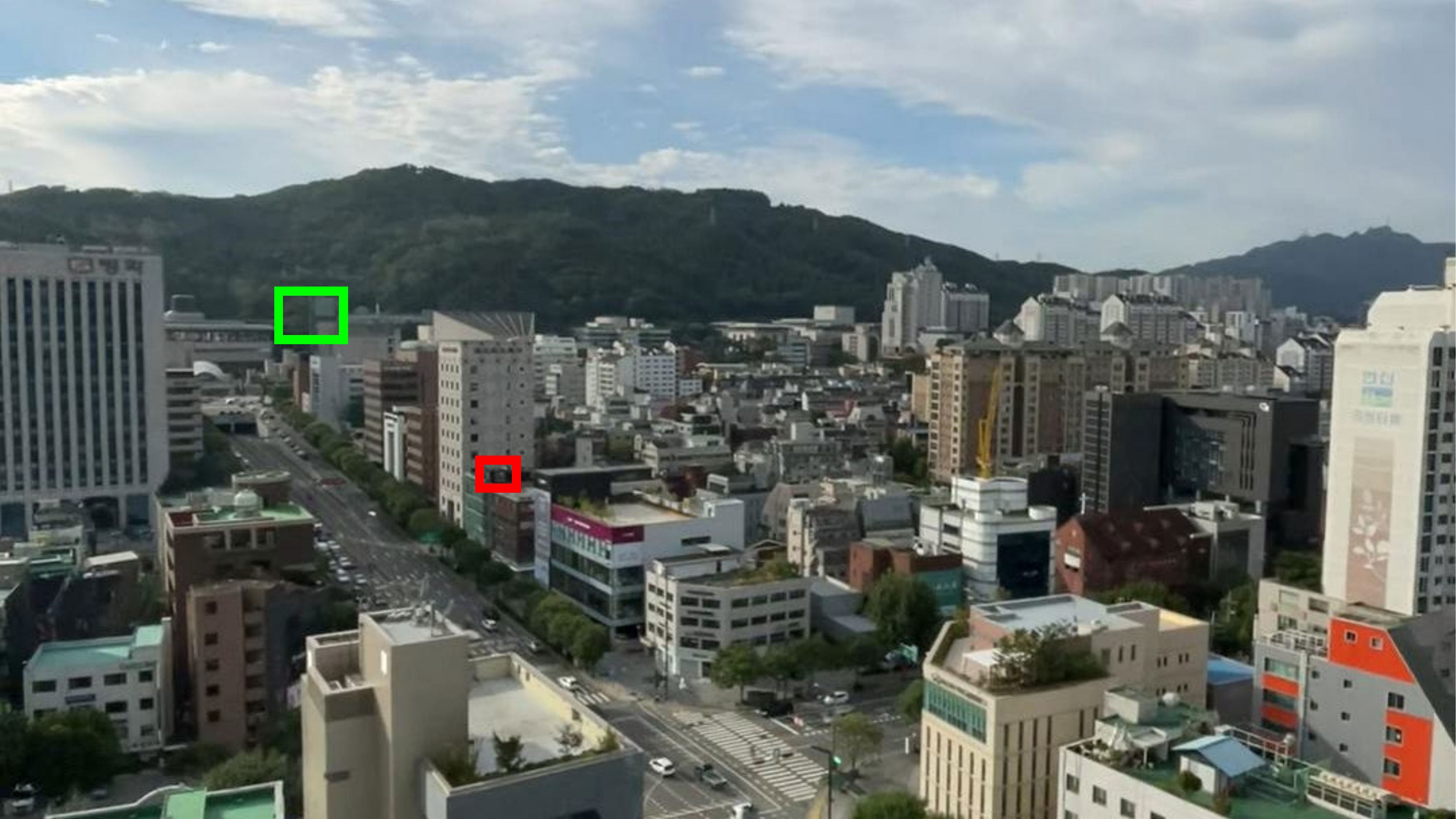}};
    \begin{scope}[x={(image.south east)},y={(image.north west)}]
        \draw[white, densely dashed, semithick] ([shift={(0.01, 0.01)}]\corlc, \corbc) rectangle ([shift={(-0.01, -0.01)}]{\corlc+\corrtc}, {\corbc+\corrtyc});
    \end{scope}
    \begin{scope}[x={(image.south east)},y={(image.north west)}]
        \coordinate (A) at (\corlc+0.01,0.01);
        \coordinate (B) at (\corlc+0.01,\corbc-0.01);
        \coordinate (C) at (\corlc+0.01,\corbc+\corrtyc+0.01);
        \coordinate (D) at (\corlc+0.01,1-0.01);
        \coordinate (E) at (\corlc+\corrtc-0.01,0.01);
        \coordinate (F) at (\corlc+\corrtc-0.01,\corbc-0.01);
        \coordinate (G) at (\corlc+\corrtc-0.01,\corbc+\corrtyc+0.01);
        \coordinate (H) at (\corlc+\corrtc-0.01,1-0.01);
        \draw[white,densely dashed,line width=0.005mm] (A) -- (B);
        \draw[white,densely dashed,line width=0.005mm] (C) -- (D);
        \draw[white,densely dashed,line width=0.005mm] (E) -- (F);
        \draw[white,densely dashed,line width=0.005mm] (G) -- (H);
    \end{scope}
    \end{tikzpicture} 
    &
    \includegraphics[width=\ws\textwidth,cfbox=red 1pt -1pt 0pt]{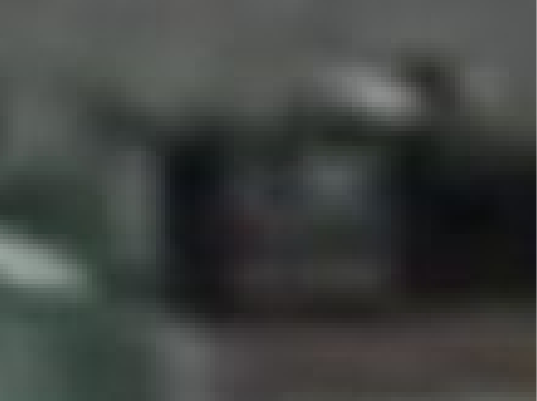}
    &
    \includegraphics[width=\ws\textwidth,cfbox=red 1pt -1pt 0pt]{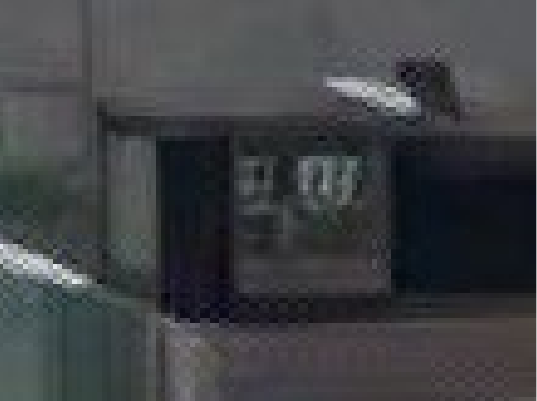}
    &
    \includegraphics[width=\ws\textwidth,cfbox=red 1pt -1pt 0pt]{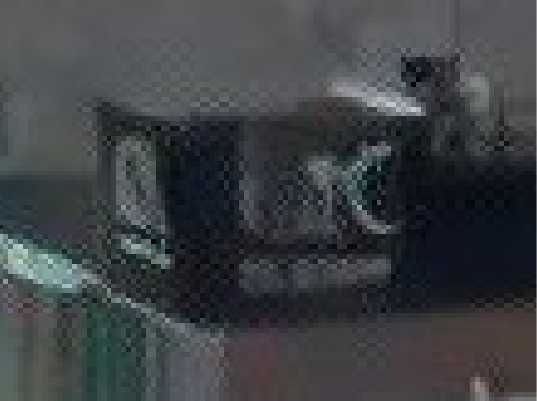}
    &
    \includegraphics[width=\ws\textwidth,cfbox=red 1pt -1pt 0pt]{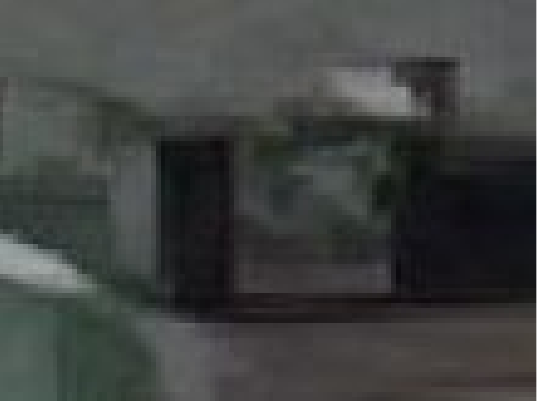}
    &
    \includegraphics[width=\ws\textwidth,cfbox=red 1pt -1pt 0pt]{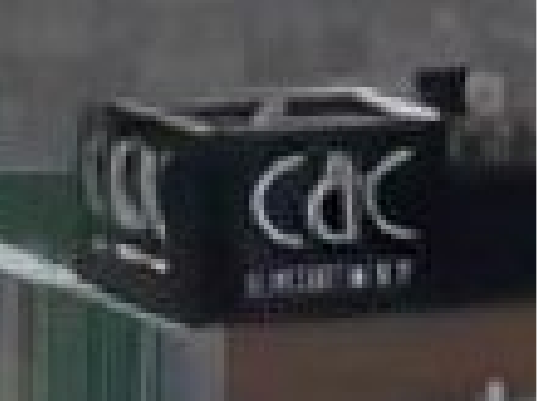}
    \\[-0.005in]
    \begin{tikzpicture}
    \node[anchor=south west,inner sep=0] (image) at (0,0) {\includegraphics[trim=0 0 0 0, clip,  width=\wb\textwidth]{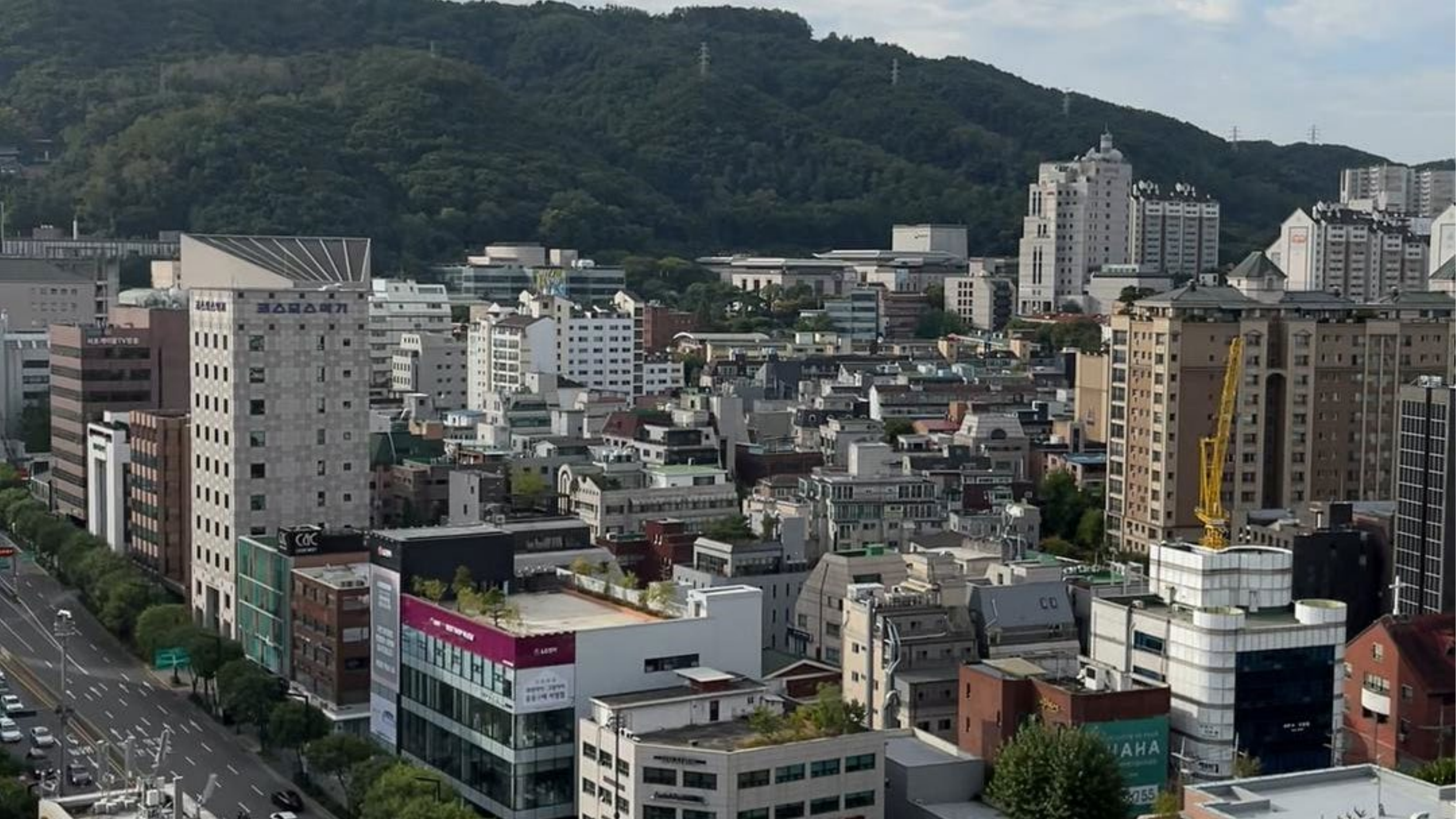}};
    \begin{scope}[x={(image.south east)},y={(image.north west)}]                                                 
        \draw[white, densely dashed, thick] ([shift={(0.007, 0.012)}]0, 0) rectangle ([shift={(-0.007, -0.012)}]{1.0}, {1.0});
    \end{scope}
    \end{tikzpicture}
    &
    \includegraphics[width=\ws\textwidth,cfbox=green 1pt -1pt 0pt]{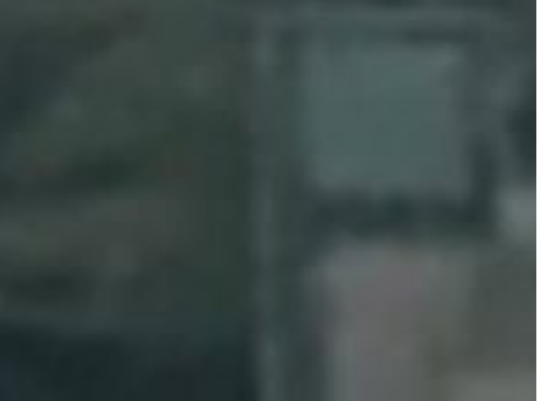}
    &
    \includegraphics[width=\ws\textwidth,cfbox=green 1pt -1pt 0pt]{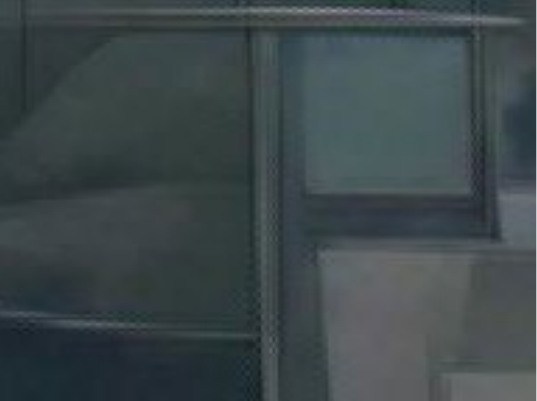}
    &
    \includegraphics[width=\ws\textwidth,cfbox=green 1pt -1pt 0pt]{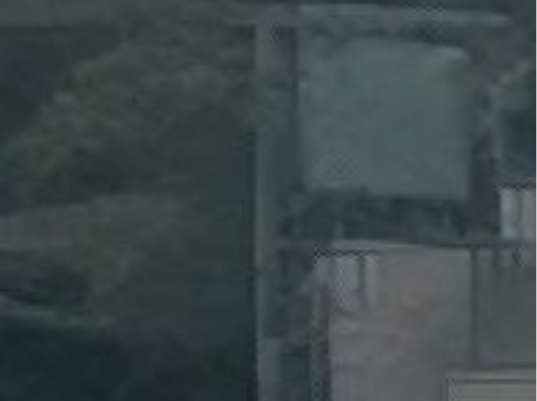}
    &
    \includegraphics[width=\ws\textwidth,cfbox=green 1pt -1pt 0pt]{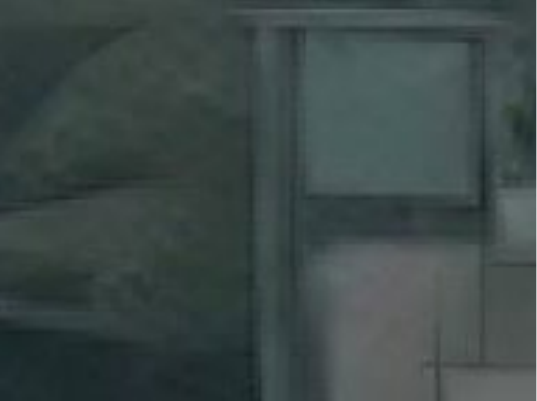}
    &
    \includegraphics[width=\ws\textwidth,cfbox=green 1pt -1pt 0pt]{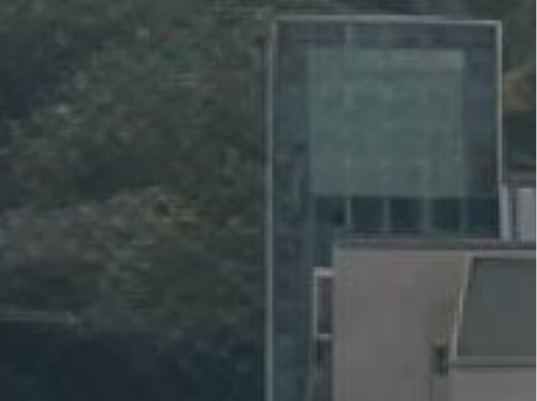}
    \\[-0.01in]
    \midrule[0.26mm]
    \end{tabular}
    
    \def \img {0074/0000}
    \def \imgr {0074/0000_red}
    \def \imgg {0074/0000_green}
    
    \footnotesize
    \begin{tabular}{cccccc}

    \\[-0.131in]

    \begin{tikzpicture}
    \node[anchor=south west,inner sep=0] (image) at (0,0) {\includegraphics[width=\wb\textwidth]{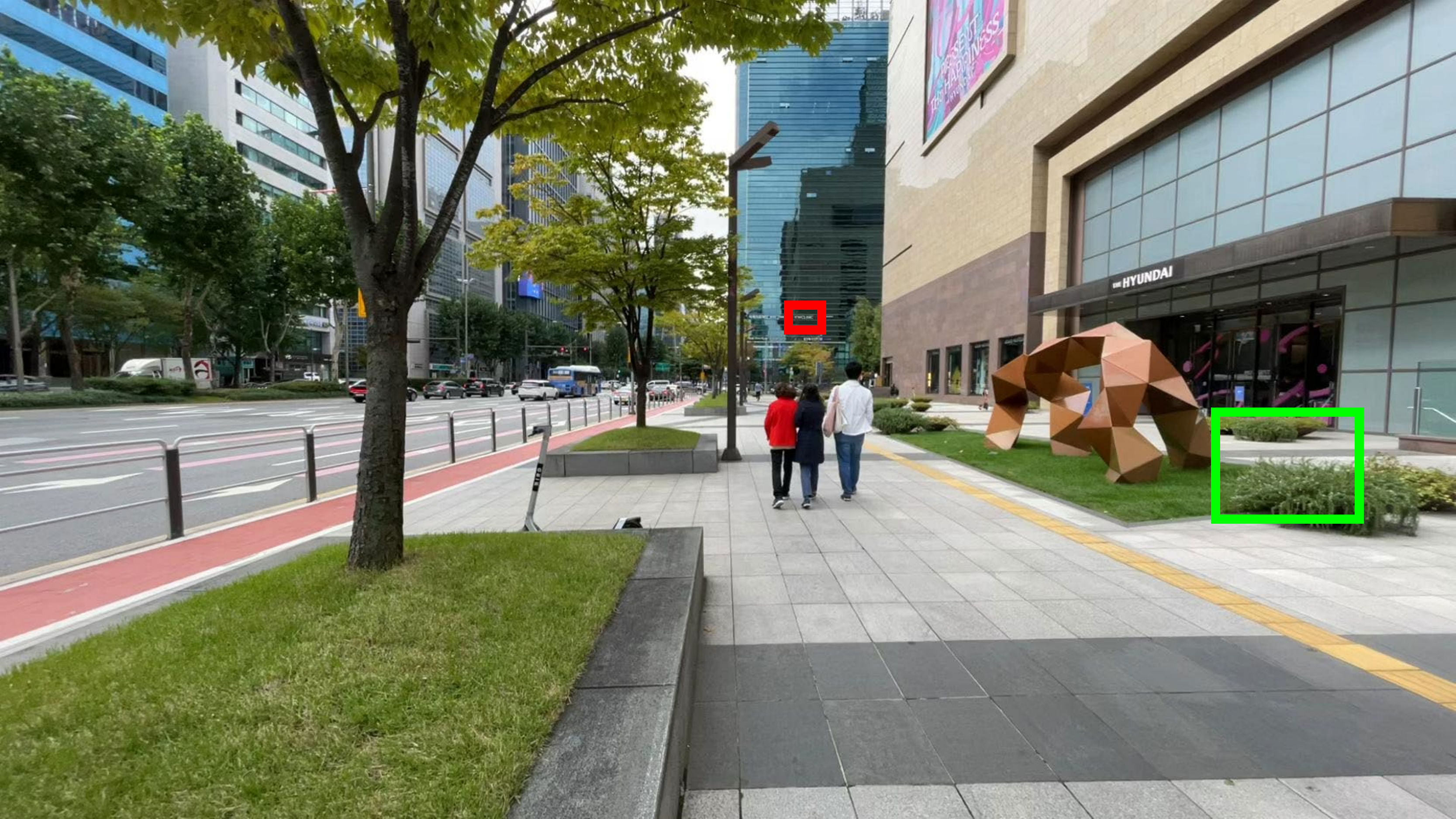}};
    \begin{scope}[x={(image.south east)},y={(image.north west)}]
        \draw[white, densely dashed, semithick] ([shift={(0.01, 0.01)}]\corlc, \corbc) rectangle ([shift={(-0.01, -0.01)}]{\corlc+\corrtc}, {\corbc+\corrtyc});
    \end{scope}
    \end{tikzpicture} 
    &
    \includegraphics[width=\ws\textwidth,cfbox=red 1pt -1pt 0pt]{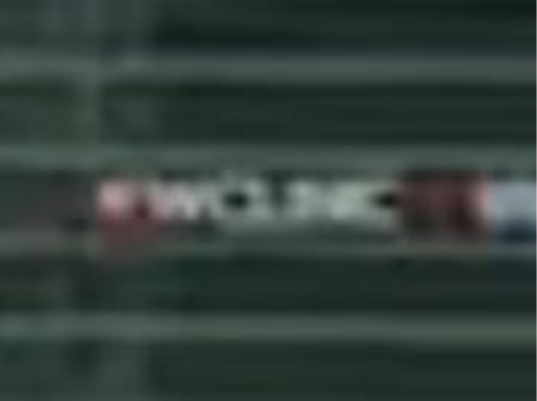}
    &
    \includegraphics[width=\ws\textwidth,cfbox=red 1pt -1pt 0pt]{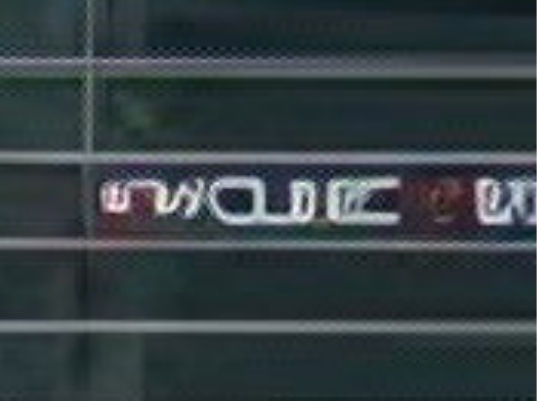}
    &
    \includegraphics[width=\ws\textwidth,cfbox=red 1pt -1pt 0pt]{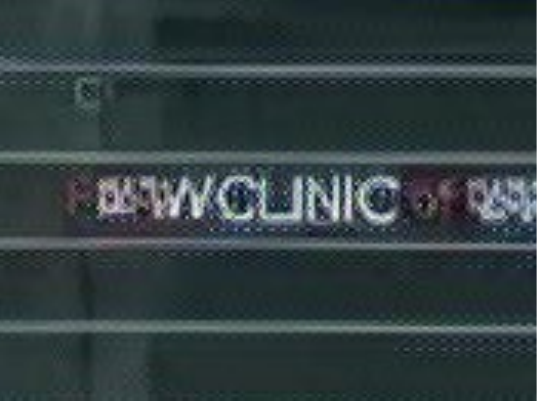}
    &
    \includegraphics[width=\ws\textwidth,cfbox=red 1pt -1pt 0pt]{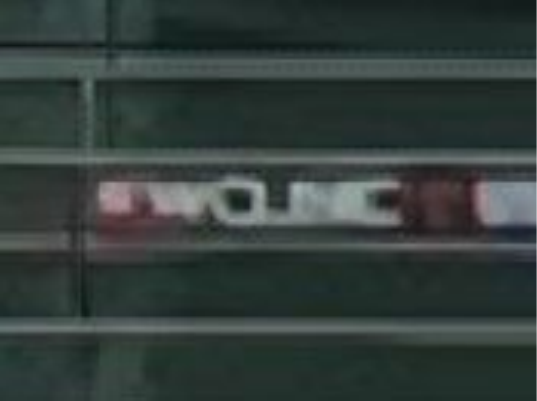}
    &
    \includegraphics[width=\ws\textwidth,cfbox=red 1pt -1pt 0pt]{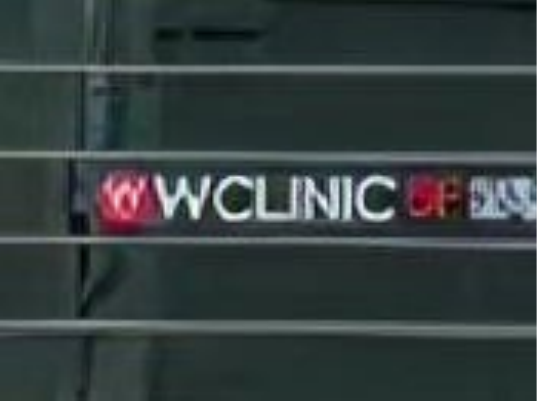}
    
    \\[-0.005in]
    
    \begin{tikzpicture}
    \node[anchor=south west,inner sep=0] (image) at (0,0) {\includegraphics[trim=0 0 0 0, clip,  width=\wb\textwidth]{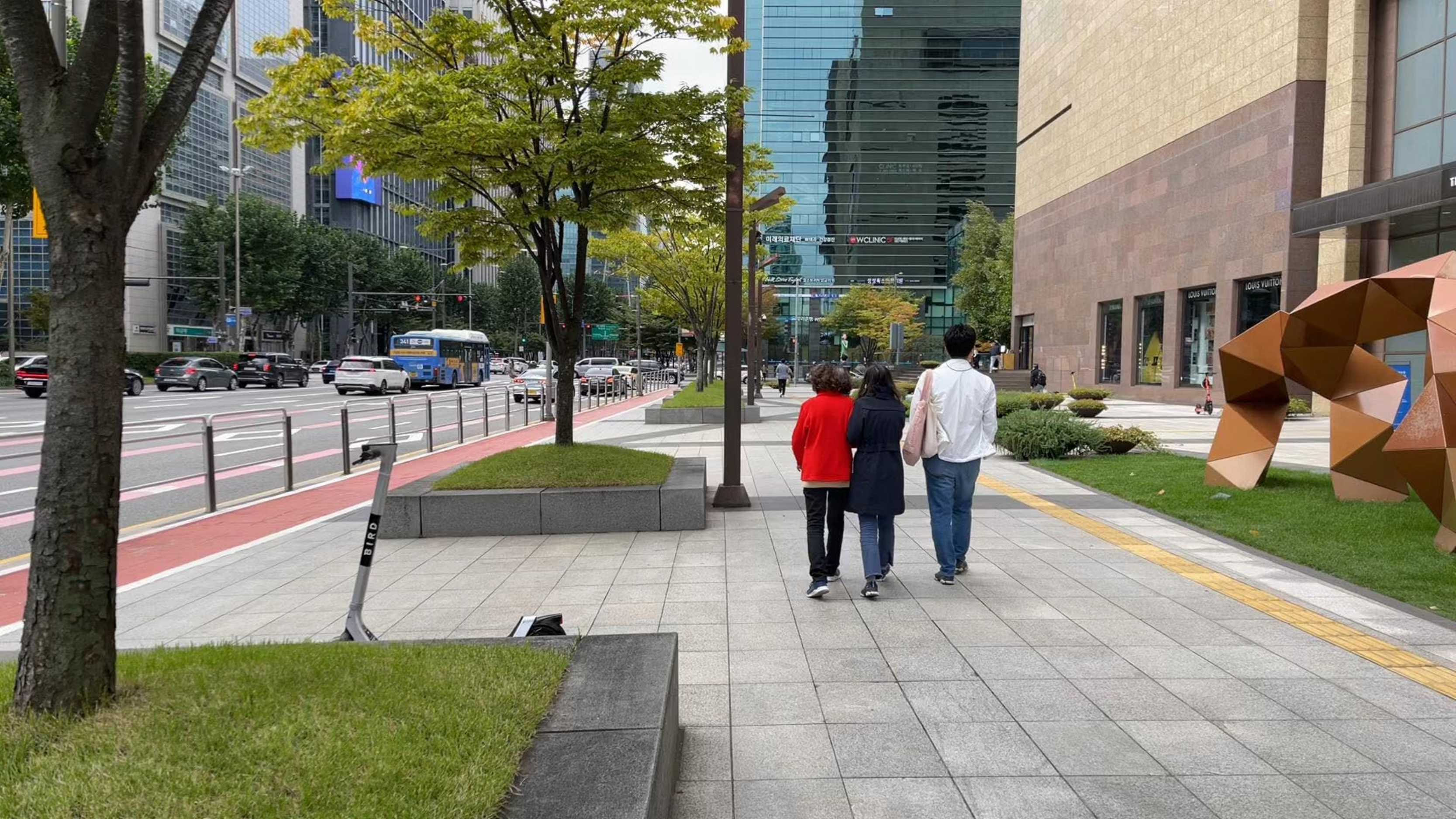}};
    \begin{scope}[x={(image.south east)},y={(image.north west)}]                                                 
        \draw[white, densely dashed, thick] ([shift={(0.007, 0.012)}]0, 0) rectangle ([shift={(-0.007, -0.012)}]{1.0}, {1.0});
    \end{scope}
    \end{tikzpicture}
    &
    
    \includegraphics[width=\ws\textwidth,cfbox=green 1pt -1pt 0pt]{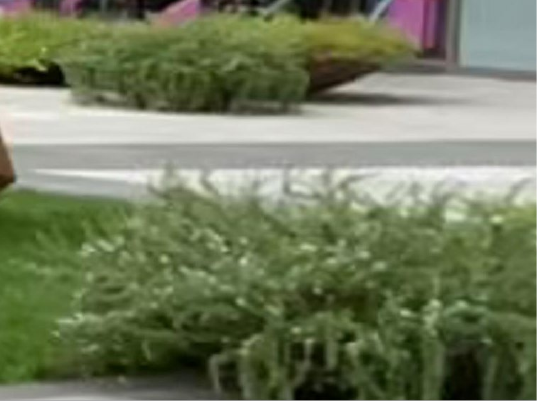}
    &
    \includegraphics[width=\ws\textwidth,cfbox=green 1pt -1pt 0pt]{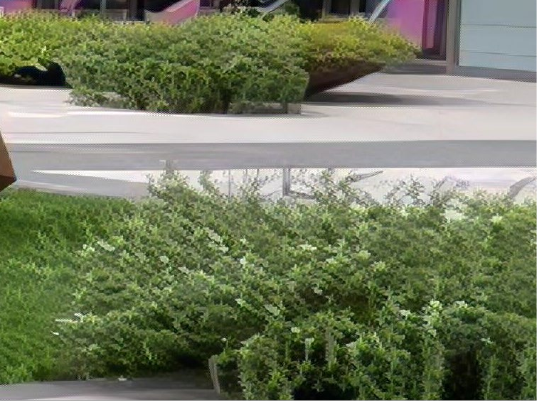}
    &
    \includegraphics[width=\ws\textwidth,cfbox=green 1pt -1pt 0pt]{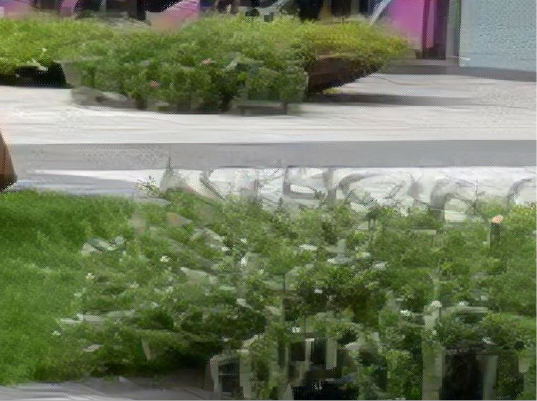}
    &
    \includegraphics[width=\ws\textwidth,cfbox=green 1pt -1pt 0pt]{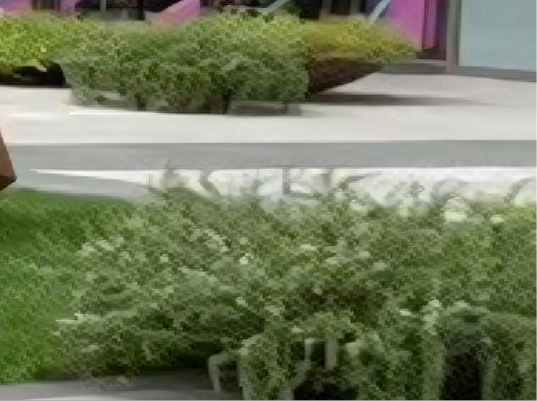}
    &
    \includegraphics[width=\ws\textwidth,cfbox=green 1pt -1pt 0pt]{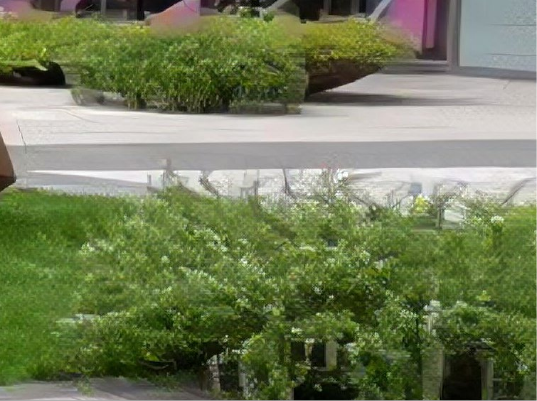}
    \\
    \end{tabular}

\vspace{-0.4cm}
\caption{Qualitative comparison on 8K 4$\times$SR video results from real-world HD videos.
}
\label{fig:comparison:8K}
\vspace{-12pt}
\end{figure*}
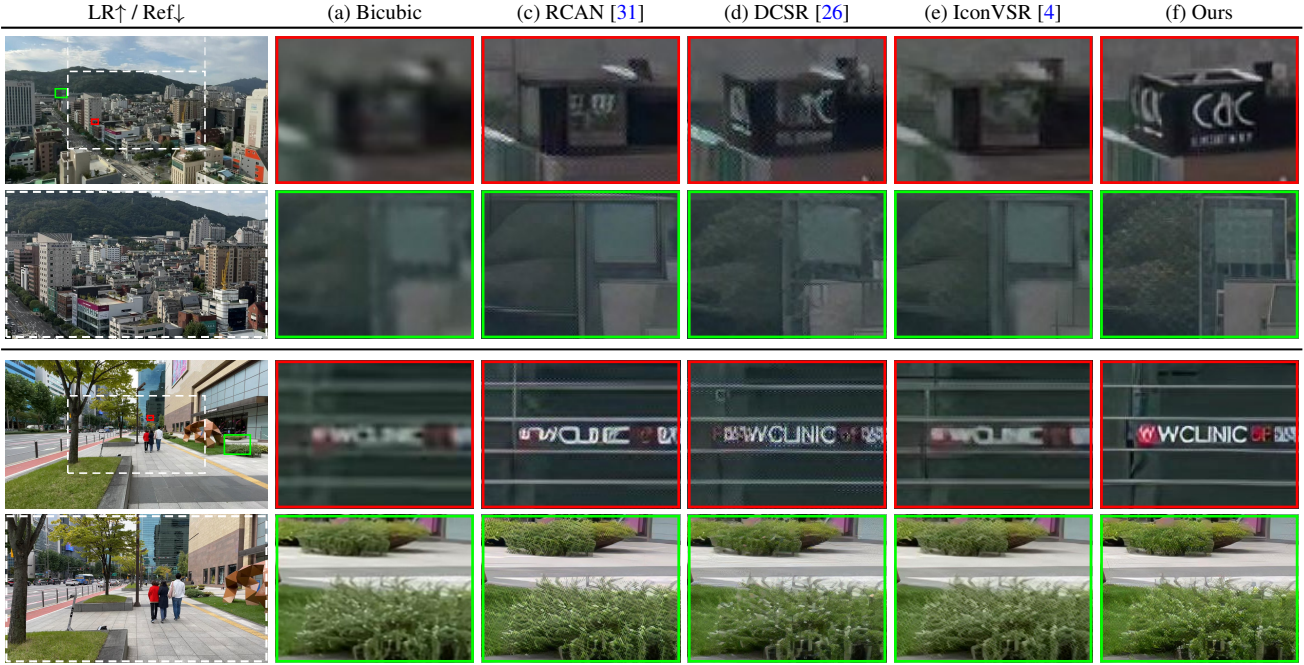

\vspace{-12pt}
\paragraph{Qualitative Comparison}
For the qualitative comparison, we show 8K ($\text{4320}\!\times\!\text{7280}$) 4$\times$SR video results given real-world HD ($\text{1080}\!\times\!\text{1920}$) videos.
For the comparison, we select the best models from each SISR, RefSR, and VSR approaches: RCAN~\cite{zhang2018rcan}, DCSR~\cite{wang2021DCSR}, and IconVSR~\cite{chan2021basicvsr}, respectively, according to their quantitative performance shown with the RealMCVSR test set.
We train each model with the proposed training strategy (\cref{sec:Real8K}).

\cref{fig:comparison:8K} shows a qualitative comparison for 8K 4$\times$SR results from real-world HD videos.
The results show that non-reference-based SR methods, RCAN and IconVSR, tend to over-exaggerate textures, while non-textured regions tend to be overly smoothed out.
The RefSR method, DCSR, shows better fidelity than RCAN and IconVSR in the overlapped FoV (red box).
However, DCSR tends to smooth out regions outside the overlapped FoV (green box).
Our method shows the best result compared to the previous ones.
Compared to DCSR, our model robustly reconstructs finer details with balanced fidelity between regions inside and outside the overlapped FoV.
Moreover, the details and textures reconstructed outside the FoV are more photo-realistic.

\vspace{-2pt}
\section{Conclusion}
We proposed the first RefVSR framework with the practical focus on videos captured in an asymmetric multi-camera setting.
To efficiently utilize a Ref video sequence, we adopted a bidirectional recurrent framework 
and proposed the propagative temporal fusion module to fuse and propagate Ref features well-matched to LR features.
To train and validate the network, we provided the RealMCVSR dataset consisting of real-world HD video triplets.
An adaptation training strategy is proposed to fully utilize video triplets in the dataset.
In the experiments, we verified the effects of key components in our model, and our model achieves the state-of-the-art 4$\times$VSR performance.

\vspace{-14pt}
\paragraph{Limitation}
As previous RefSR methods~\cite{Zhang2019ImageSB,yang2020TTSR,Xie2020FeatureRM,wang2021DCSR}, our network consumes quite an amount of memory for applying global matching between real-world HD frames.
We plan to develop a memory-efficient RefVSR framework.

\vspace{-14pt}
\small
\paragraph{\small Acknowledgments} %
We thank Hyeongseok Son for helpful discussions
and Jihye Kim and Anna Choi for their help in collecting the RealMCVSR dataset.
This work was supported by the Ministry of Science and ICT, Korea,
through 
IITP grants
(SW Star Lab, 2015-0-00174;
AI Innovation Hub, 2021-0-02068;
Artificial Intelligence Graduate School Program (POSTECH), 2019-0-01906)
and
NRF grants (2018R1A5A1060031; 2020R1C1C1014863).

{\small
\bibliographystyle{compile/ieee_fullname}
\bibliography{ms}
}
\end{document}